\documentclass{article}

 \usepackage[main, final]{neurips_2026}

\usepackage[utf8]{inputenc} 
\usepackage[T1]{fontenc}    
\usepackage{hyperref}       
\usepackage{url}            
\usepackage{booktabs}       
\usepackage{amsfonts}       
\usepackage{nicefrac}       
\usepackage{microtype}      
\usepackage{xcolor}         

\usepackage{graphicx}
\usepackage{amsmath}
\usepackage{amssymb}
\usepackage{mathtools}
\usepackage{amsthm}
\usepackage{multirow}
\usepackage[table]{xcolor}  
\usepackage{pifont}
\usepackage{algorithm}
\usepackage{algorithmic}
\usepackage{wrapfig}
\usepackage{makecell}

\title{Towards Explainable Industrial Anomaly Detection via Knowledge-Guided Latent Reasoning}

\author{
\textbf{Peng Chen}$^{1}$ \quad
\textbf{Wenqiang Wang}$^{1}$ \quad
\textbf{Yunkang Cao}$^{2}$ \quad
\textbf{Chengliang Liu}$^{3}$ \\[3pt]
\textbf{Wei Wang}$^{1}$ \quad
\textbf{Mingbo Yang}$^{1}$ \quad
\textbf{Li Shen}$^{1}$ \quad
\textbf{Wenqi Ren}$^{1}$ \\[3pt]
\textbf{Xiaochun Cao}$^{1}$ \quad
\textbf{Chao Huang}$^{1}$%
\thanks{Corresponding author.}
\\[6pt]
{\small
$^{1}$Shenzhen Campus of Sun Yat-sen University}
{\small
$^{2}$Hunan University}
{\small
$^{3}$University of Macau}\\[5pt]
{\small
\texttt{\{chenp299, wangwq69, yangmb3\}@mail2.sysu.edu.cn}}\\ 
{\small
\texttt{liucl1996@163.com}},
{\small
\texttt{caoyunkang0207@gmail.com}}\\
{\small
\texttt{\{wangwei29, shenli6, renwq3, caoxiaochun, huangch253\}@mail.sysu.edu.cn}}
}

\begin{document}

\maketitle

\begin{abstract}
Industrial anomaly detection demands precise reasoning over fine-grained defect patterns. However, existing multimodal large language models (MLLMs), pretrained on general-domain data, often struggle to capture category-specific anomalies, thereby limiting both detection accuracy and interpretability. To address these limitations, we propose Reason-IAD, a knowledge-guided dynamic latent reasoning framework for explainable industrial anomaly detection. Reason-IAD comprises two core components. First, a retrieval-augmented knowledge module incorporates category-specific textual descriptions into the model input, enabling context-aware reasoning over domain-specific defects. Second, an entropy-driven latent reasoning mechanism conducts iterative exploration within a compact latent space using optimizable latent think tokens, guided by an entropy-based reward that encourages confident and stable predictions. Furthermore, a dynamic visual injection strategy selectively incorporates the most informative image patches into the latent sequence, directing the reasoning process toward regions critical for anomaly detection. Extensive experimental results demonstrate that Reason-IAD consistently outperforms state-of-the-art methods across multiple tasks. The code will be publicly available at~\url{https://github.com/chenpeng052/Reason-IAD}.
\end{abstract}

\section{Introduction}
 
Industrial anomaly detection plays a critical role in ensuring the safety and reliability of modern manufacturing systems~\cite{jiang2024mmad,zhang2025towards,fangdemeaned}, aiming to accurately identify anomalous samples that deviate from normal patterns in complex visual data~\cite{ma2025aa,wang2025cnc,chen2026dyc}. However, conventional methods typically rely on carefully curated, task-specific datasets, rendering them highly sensitive to domain priors and limiting their generalization across diverse scenarios and product categories~\cite{cao2024adaclip,qu2025bayesian,wang2025decad}. In practice, substantial distribution discrepancies across different industrial products and imaging conditions introduce severe domain shifts, hindering their deployment in real-world manufacturing environments.
\begin{figure}
    \centering
    \includegraphics[width=\linewidth]{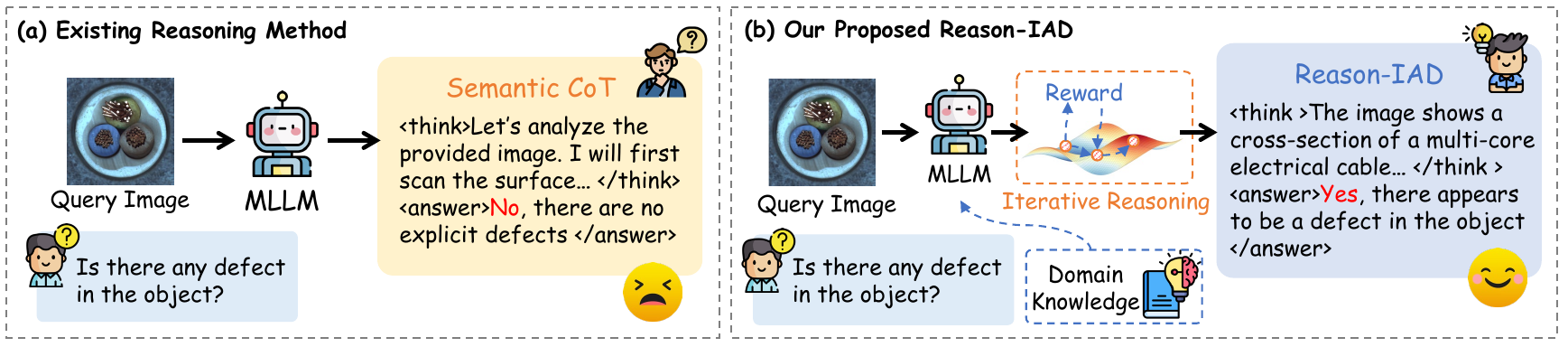}
    \caption{Comparison between existing reasoning methods and the proposed Reason-IAD. (a) Existing methods conduct reasoning through explicit chains of thought. (b) Reason-IAD retrieves domain-specific knowledge and identifies anomalies via iterative latent reasoning.}
    \label{fig:intro}
\end{figure}

In recent years, vision-language models have demonstrated considerable potential for industrial anomaly detection due to their strong multimodal representation and generalization capabilities~\cite{qu2024vcp,huang2026noise,cai2026hepe++}. CLIP-based approaches align visual features with textual semantics, enabling anomaly detection under zero-shot and few-shot settings~\cite{yan2025wavelet,cai2025towards}. For instance, AnomalyCLIP~\cite{zhou2023anomalyclip} introduces category-agnostic textual embeddings to reduce reliance on manually designed priors, while AdaCLIP~\cite{cao2024adaclip} and BayPFL~\cite{qu2025bayesian} replace static prompts with learnable textual embeddings to enhance semantic expressiveness and model adaptability. Despite these advances, the performance of such methods remains constrained by the representational capacity of pretrained models~\cite{jeong2023winclip,li2024promptad,wang2025cnc}. When confronted with complex industrial scenes and fine-grained defect patterns, they often fail to achieve reliable anomaly recognition~\cite{gu2024filo,qu2024vcp,wang2025icc,huang2024multimodal}. Moreover, these approaches generally lack explicit reasoning mechanisms, preventing evidence-based analysis as human experts would do, thus limiting their applicability in high-reliability industrial scenarios.

Recent advances in multimodal large language models (MLLMs) have opened a promising avenue for developing general-purpose, expert-level industrial anomaly detection systems~\cite{gu2024anomalygpt,li2025iad,xu2026osagen}, owing to their strong pretraining and reasoning capabilities~\cite{yan2025wavelet,tan2025reason,wang2026task}. MMAD~\cite{jiang2024mmad} introduced the first systematic benchmark for anomaly detection reasoning, evaluating models’ reasoning abilities across multiple dimensions. Building on this, Anomaly-R1~\cite{chao2025anomalyr1} leverages the GRPO training strategy to construct an end-to-end anomaly detection framework that replaces traditional expert systems, while AD-FM~\cite{liao2025ad} adopts a two-stage reasoning paradigm, combining supervised fine-tuning (SFT) with GRPO to progressively adapt pretrained multimodal models to anomaly detection tasks. However, as illustrated in Figure~\ref{fig:intro} (a), these approaches typically rely on explicitly generating long reasoning chains (chain-of-thought) to reach decisions, resulting in substantial inefficiency and limited scalability~\cite{sun2025latent,li2025latent,wang2026incomplete}. Furthermore, since the underlying models are primarily pretrained on general-domain data, they often lack task-specific domain knowledge for industrial anomaly detection, which constrains the reliability and stability of their reasoning outcomes.

To address these challenges, we propose Reason-IAD, a training-free framework for explainable industrial anomaly detection that leverages knowledge-guided dynamic latent reasoning. As illustrated in Figure~\ref{fig:intro}(b), Reason-IAD consists of two key components. First, a retrieval-augmented knowledge module incorporates category-specific textual descriptions into the model input, enabling context-aware reasoning over domain-specific defects. Second, an entropy-driven latent reasoning mechanism conducts iterative exploration within a compact latent space using optimizable latent think tokens, facilitating confident and stable reasoning. In addition, a dynamic visual injection strategy selectively injects the most informative image regions into the latent sequence, guiding the reasoning process toward critical visual evidence for anomaly detection. We evaluate Reason-IAD across seven sub-tasks under one-shot and zero-shot settings. Extensive experiments demonstrate that Reason-IAD consistently outperforms existing methods, highlighting its effectiveness and interpretability.

Our main contributions are summarized as follows:
\begin{itemize}
    \item To the best of our knowledge, this is the first work to leverage MLLMs for training-free reasoning in the context of industrial anomaly detection.
    \item We propose Reason-IAD, a knowledge-guided dynamic latent reasoning framework for explainable industrial anomaly detection, enabling effective anomaly perception through iterative reasoning guided by visual evidence and entropy-driven reward optimization.
    \item Extensive experiments demonstrate the effectiveness and generality of Reason-IAD, providing an efficient and practical solution for applying MLLMs to industrial anomaly detection.
\end{itemize}

\section{Related Work}

\subsection{Anomaly Detection}
Anomaly detection aims to identify abnormal patterns that deviate from the normal data distribution~\cite{liao2025ad, cai2025botwavenet,wang2026unifying}. However, substantial variability in appearance and the scarcity of anomalous samples make anomaly detection a particularly challenging task~\cite{cai2024multiscale,chen2026vfad,fengomiad}. Traditional unsupervised methods learn representations solely from normal samples and identify anomalies during testing~\cite{roth2022towards, deng2022anomaly,zhang2025wave}. While effective to some extent, these approaches often rely on category-specific models, which constrain their generalization and practical applicability in real-world industrial scenarios. Recent studies have explored vision–language models for zero-shot and few-shot anomaly detection by aligning visual representations with textual semantics~\cite{qu2024vcp,xu2025towards,chen2026wmoe}. For example, WinCLIP~\cite{jeong2023winclip} leverages CLIP~\cite{radford2021learning} to measure similarity between image patch features and hand-crafted text prompts. To improve cross-modal alignment, AdaCLIP~\cite{cao2024adaclip} and BayPFL~\cite{qu2025bayesian} further incorporate image information into the prompt construction process, enhancing the model’s capacity to capture anomaly-related semantics. Despite the performance gains of these approaches, their inference pipelines predominantly rely on static similarity matching, and the limited reasoning ability of base models remains a key bottleneck.

\subsection{Visual Reasoning}
Visual reasoning is a fundamental challenge in artificial general intelligence, requiring models to perform complex cognitive reasoning grounded in visual perception~\cite{yang2026beyond,wang2025vl}. This capability is essential for a wide range of tasks, including visual counting, geometric problem solving, and anomaly detection~\cite{shi2025search, chen2025synergistic, wan2025srpo}. Early approaches primarily relied on program synthesis or neuro-symbolic reasoning frameworks~\cite{amizadeh2020neuro,gupta2023visual}. Recent advances in vision–language models have enabled the integration of large language models to enhance visual reasoning. Representative works such as Reason-RFT~\cite{tan2025reason} adopt a two-stage training strategy combining supervised fine-tuning and reinforcement learning to improve visual reasoning. In anomaly detection, Anomaly-R1~\cite{chao2025anomalyr1} introduces GRPO-based reasoning alignment, while AD-FM~\cite{liao2025ad} bridges anomaly perception and explanation. Triad~\cite{li2025triad} further incorporates expert-guided region awareness with chain-of-thought reasoning to improve defect understanding. However, most existing approaches rely heavily on large-scale reconstructed datasets for fine-tuning, resulting in high training costs and limited scalability in real-world applications.

\subsection{Latent Reasoning}
Recent research on advanced reasoning in multimodal large language models has evolved from explicit text-based reasoning toward more efficient latent space reasoning approaches~\cite{sun2025latent,tan2025think,li2025seek}. In latent space reasoning, the intermediate “thinking process” is encoded as continuous vectors in the model’s hidden space rather than as discrete textual representations, aiming to achieve performance comparable to explicit chain-of-thought (CoT) reasoning while using a more compact and efficient representation~\cite{chen2025reasoning,ruan2025reasoning,geiping2025scaling}. Some methods enhance latent representations through carefully designed training strategies to enable more effective interactive reasoning~\cite{hao2024training, yang2025machine,huang2025thinkact}, while others adopt training-free approaches that guide latent activations to refine the reasoning process~\cite{zhang2025soft,liu2025reasoning,pham2025multimodal}. However, existing methods generally lack knowledge-guided reasoning and the integration of visual context, limiting their effectiveness on tasks that require external knowledge or visual evidence. In contrast, our approach retrieves relevant knowledge and dynamically integrates visual evidence into the latent reasoning process, improving performance in industrial anomaly detection.

\section{Methodology}
\begin{figure*}
    \centering
    \includegraphics[width=\linewidth]{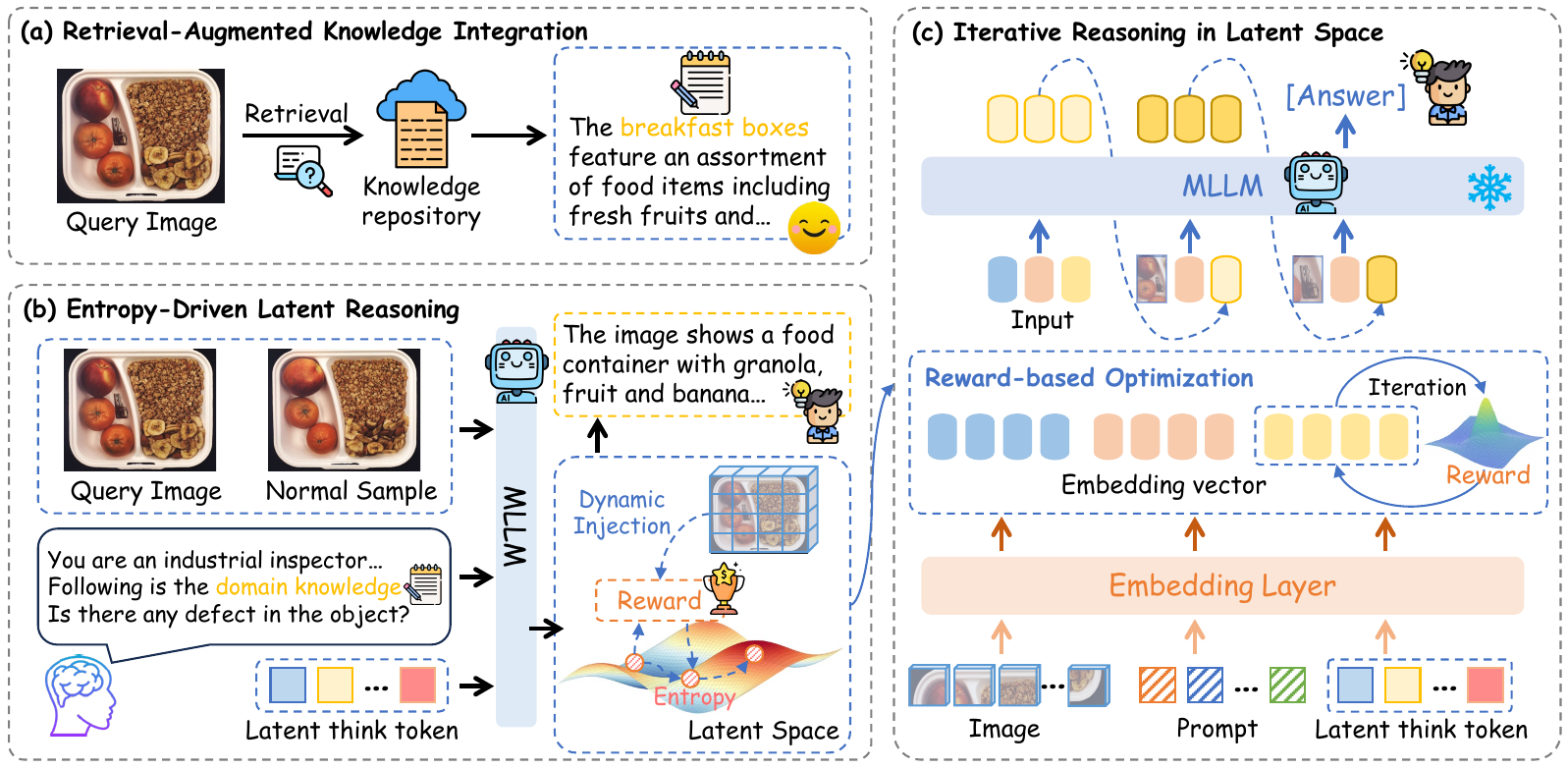}
    \caption{Overview of the proposed Reason-IAD. (a) Given a query image, Reason-IAD retrieves the most relevant category-specific descriptions and incorporates them into the model prompt to enhance anomaly awareness. (b) An entropy-guided latent reasoning module iteratively refines latent think tokens and dynamically injects visual evidence to improve reasoning accuracy. (c) Illustration of the iterative latent-space reasoning process, in which reward signals and visual cues progressively guide the model toward the final prediction.}
    \label{fig:method}
\end{figure*}

\subsection{Problem Definition}
In this work, we study a multimodal anomaly detection question-answering (QA) task, in which the model answers questions based on a given industrial image.

Formally, given a query image $I\in \mathcal{I}_{test}$ depicting an industrial object and a question $Q\in \mathcal{Q}_{test}$, the model aims to predict the correct answer $A$. Each question corresponds to a specific anomaly reasoning objective, demanding varying levels of visual perception and reasoning.

Our proposed method supports two settings: (1) \textbf{One-shot setting}: the model receives a query image $I$ and a question $Q$, along with a reference image $I_{r} \in \mathcal{I}_{ref}$ of a normal object from the same domain. The model is required to generate both the answer and a corresponding explanation.
(2) \textbf{Zero-shot setting}: the model only receives the query image $I$ and the question $Q$, without any reference information, and is expected to produce the answer along with the relevant explanation.

\subsection{Reason-IAD}
In this paper, we propose Reason-IAD, a knowledge-guided dynamic latent reasoning framework for explainable industrial anomaly detection. As illustrated in Figure~\ref{fig:method}, Reason-IAD comprises two principal components: (1) a Retrieval-Augmented Knowledge Integration (RAKI) module that retrieves category-specific textual descriptions and incorporates them into the model prompt, enabling context-aware reasoning over domain-specific defects; and (2) an Entropy-Driven Latent Reasoning mechanism (EDLR) that performs iterative exploration in a compact latent space using optimizable latent think tokens. Furthermore, inspired by human cognitive behavior in revisiting visual evidence under uncertainty, we introduce a dynamic visual injection strategy that selectively incorporates the most informative image patches into the latent sequence, guiding the reasoning process toward regions that are critical for anomaly detection.

\subsection{Retrieval-Augmented Knowledge Integration}
Existing multimodal large language models are predominantly pretrained on general-domain data, which often limits their ability to understand industrial anomalies. To improve both reasoning accuracy and explainability in industrial anomaly detection tasks, we propose a retrieval-augmented mechanism that incorporates category-specific knowledge to enrich the model input prompt.

Specifically, given a query image $I$ and its associated question $Q$, we construct a category-oriented knowledge repository $\mathcal{K} = \{(c_i, \text{desc}_i)\}_{i=1}^{N}$, where $c_i$ denotes the $i$-th category label (e.g., \texttt{cable}, \texttt{breakfast box}), and $\text{desc}_i$ provides a detailed textual description of the category, encompassing both normal characteristics and defect patterns. This repository serves as an external knowledge source for contextual augmentation.   
To identify the most relevant category information, the query image is first encoded into a visual embedding $\mathbf{v} \in \mathbb{R}^{d}$ using a visual encoder $f_v$, while each category label $c_i$ is encoded into a textual embedding $\mathbf{k}_i \in \mathbb{R}^{d}$ using a text encoder $f_t$, projecting both modalities into a shared semantic space:
\begin{align}
    \mathbf{v} = f_v(I), \quad
    \mathbf{k}_i = f_t(c_i).
\end{align}

We then evaluate the semantic alignment between the query image and each category label by computing their cosine similarity:
\begin{align}
    s_i = \frac{\mathbf{v}^\top \mathbf{k}_i}{\|\mathbf{v}\| \, \|\mathbf{k}_i\|}, \quad i = 1, \dots, N,
\end{align}
where $s_i$ denotes the similarity score between the query image and the $i$-th category.
According to the similarity scores $\{s_i\}$, we retrieve the top-$k$ categories with the highest scores. The corresponding category descriptions are then incorporated into the prompt, forming an enriched contextual representation that provides category-aware guidance for reasoning and explanation generation.

\subsection{Entropy-Driven Latent Reasoning}
Industrial anomaly detection often demands sophisticated reasoning over subtle visual patterns and contextual knowledge. Traditional approaches, such as Chain-of-Thought (CoT) prompting, rely on generating explicit textual reasoning steps to guide model predictions, but they typically entail lengthy intermediate reasoning processes. Inspired by the intertwined nature of perception and reasoning in human cognition, we propose Reason-IAD, an entropy-driven latent reasoning framework. Within this framework, the model dynamically integrates visual evidence from the query image with category-specific knowledge from an external repository in the latent space, producing a compact and interpretable reasoning representation that enables robust industrial anomaly detection.

\textbf{Latent Think Tokens.} 
To facilitate dynamic exploration in the latent space, we introduce a set of latent think tokens, denoted as $\mathcal{Z} = \{z_1, z_2, \dots, z_m\}$, which are initialized as optimizable vector embeddings and inserted into the original multimodal input sequence. During each iteration of latent reasoning, we incorporate controlled stochasticity to expand the search space while preserving the stability of the representations. Specifically, we adopt additive Gaussian noise as a local perturbation strategy. Given the latent think tokens $\mathcal{Z}^{(n)}$ at iteration $n$, the perturbed state is defined as:
\begin{equation}
\label{eq:noise_addition}
\mathcal{Z}'^{(n)} = \mathcal{Z}^{(n)} + \xi^{(n)}, \quad \xi^{(n)} \sim \mathcal{N}(0, \sigma^2 I),
\end{equation}
where $\sigma^2$ is the variance of the Gaussian noise, which controls the perturbation magnitude, and $\xi^{(n)}$ represents the Gaussian noise sampled at iteration $n$ to encourage diverse latent trajectories. This strategy introduces controlled exploration in the latent space while preserving the semantic consistency of the original representations, facilitating more effective and robust latent reasoning.



\textbf{Entropy-Guided Reward Optimization.}
In multimodal anomaly detection, the objective is to identify and localize anomalous regions given a query image along with its corresponding textual prompt. Let $Q = \{q_1, \dots, q_l\}$ denote the sequence of input prompt tokens, and $I = \{i_1, \dots, i_t\}$ represent the set of visual embeddings extracted from the input image. A MLLM parameterized by $\theta$ integrates these multimodal inputs to generate a latent reasoning sequence $\mathcal{G} = \{g_1, \dots, g_N\}$ that supports anomaly detection. The generation process can be formally described using the following autoregressive factorization:
\begin{equation}
\pi_\theta(\mathcal{G} | Q, I) = \prod_{n=1}^{N} \pi_\theta(g_n | \mathcal{G}_{<n}, Q, I),
\end{equation}
where $\mathcal{G}_{<n}$ denotes the sequence of tokens generated prior to step $n$.
Accurate reasoning in industrial anomaly detection requires not only reliable predictions but also well-calibrated uncertainty over the model’s internal reasoning states. Existing latent reasoning frameworks typically lack explicit mechanisms to regulate the uncertainty of intermediate representations, potentially resulting in ambiguous or unstable predictions. To address this limitation, we propose an entropy-guided reward mechanism that leverages predictive uncertainty as an intrinsic feedback signal for optimizing latent think tokens. Specifically, for each query instance, we quantify the model’s predictive uncertainty by computing the entropy of its probability distribution as:
\begin{equation}
\label{eq:entropy}
\mathcal{H}(P_i^{(n)}) = -\sum_{j=1}^{C} P_i^{(n)}(j) \log P_i^{(n)}(j)
,
\end{equation}
where $P_i^{(n)}$ denotes the predicted probability distribution over answers by the $i$-th latent think token at iteration $n$. Lower entropy indicates that the model has reached a more confident and deterministic reasoning state, while higher entropy reflects higher epistemic uncertainty arising from unfamiliar or conflicting features. Based on this measure, the global reward at iteration $n$ is defined as the complement of the mean entropy across all latent think tokens:
\begin{equation}
\label{eq:reword}
\mathcal{R}(\mathcal{Z}^{(n)}) = 1 - \frac{1}{m} \sum_{i=1}^{m} \mathcal{H}(P_i^{(n)}),
\end{equation}
where higher reward values incentivize the model to reduce uncertainty in its latent reasoning process, thereby encouraging more stable predictions. To iteratively optimize the latent think tokens toward higher confidence (i.e., lower entropy), we adopt a REINFORCE-based policy gradient method~\cite{montenegro2024last}. Specifically, given the stochastic perturbations of latent tokens defined in Eq.~\eqref{eq:noise_addition}, the update rule at iteration $n$ is formulated as:
\begin{equation}
\label{eq:iteration}
\mathcal{Z}'^{(n)} \leftarrow \mathcal{Z}^{(n)} + \eta \nabla_{\mathcal{Z}^{(n)}} J(\mathcal{Z}^{(n)}),
\end{equation}
where $\eta$ represents the learning rate, and $J(\cdot)$ denotes the expected reward objective. 
Following the Policy Gradient Theorem and considering the Gaussian perturbations, the gradient of the objective can be expressed as:
\begin{equation}
\begin{aligned}
\nabla_{\mathcal{Z}} \mathcal{J}(\mathcal{Z}) &= \mathbb{E}_{\mathcal{Z}' \sim \pi(\cdot | \mathcal{Z})} \left[ \mathcal{R}(\mathcal{Z}') \nabla_{\mathcal{Z}} \log \pi(\mathcal{Z}' | \mathcal{Z}) \right]
= \mathbb{E} \left[ \mathcal{R}(\mathcal{Z}') \frac{\xi}{\sigma^2} \right].
\end{aligned}
\end{equation}
This optimization process encourages the latent tokens to reduce predictive uncertainty, thereby improving the overall reasoning accuracy of the model.

\begin{table*}[]
\caption{Performance comparison of commercial and open-source MLLMs on the MMAD benchmark under the standard one-shot setting. The best results are highlighted in bold, and all metrics are reported as accuracy (\%).}
\label{table:1shot}
\resizebox{\linewidth}{!}{
\begin{tabular}{ccccccccccc}
\toprule
                                                     &                         &                         & Anomaly        & \multicolumn{4}{c}{Defect}                             & \multicolumn{2}{c}{Object} &                           \\ \cline{4-10}
\multirow{-2}{*}{Type}                               & \multirow{-2}{*}{Model} & \multirow{-2}{*}{Scale} & Discrimination & Classification & Localization & Description & Analysis & Classification  & Analysis & \multirow{-2}{*}{Average} \\ \hline
Random                                               & Random Chance           & -                       & 50.00          & 25.00          & 25.00        & 25.00       & 25.00    & 25.00           & 25.00    & 28.57                     \\ \midrule
\rowcolor[HTML]{ECF4FF} 
\cellcolor[HTML]{ECF4FF}                             & Human (expert)          & -                       & 95.24          & 75.00          & 92.31        & 83.33       & 94.20    & 86.11           & 80.37    & 86.65                     \\
\rowcolor[HTML]{ECF4FF} 
\multirow{-2}{*}{\cellcolor[HTML]{ECF4FF}Human}      & Human (ordinary)        & -                       & 86.90          & 66.25          & 85.58        & 71.25       & 81.52    & 89.58           & 69.72    & 78.69                     \\ \midrule
\rowcolor[HTML]{EFEFEF} 
\cellcolor[HTML]{EFEFEF}                             & Claude-3.5-sonnet       & -                       & 60.14          & 60.14          & 48.81        & 67.13       & 79.11    & 85.19           & 79.83    & 68.62                     \\
\rowcolor[HTML]{EFEFEF} 
\cellcolor[HTML]{EFEFEF}                             & Gemini-2.5-pro          & -                       & 83.07          & \textbf{73.86}          & 67.20        & \textbf{79.97}       & 86.27    & 94.88           & 83.08    & \textbf{81.19}                     \\
\rowcolor[HTML]{EFEFEF} 
\cellcolor[HTML]{EFEFEF}                             & Gemini-2.5-flash        & -                       & \textbf{83.38}          & 69.88          & 63.30        & 76.41       & 81.57    & 94.04           & 82.00    & 78.65                     \\
\rowcolor[HTML]{EFEFEF} 
\cellcolor[HTML]{EFEFEF}                             & GPT-4o                  & -                       & 68.63          & 65.80          & 55.62        & 73.21       & 83.41    & \textbf{94.98}           & 82.80    & 74.92                     \\
\rowcolor[HTML]{EFEFEF} 
\multirow{-5}{*}{\cellcolor[HTML]{EFEFEF}Commercial} & GPT-5-mini              & -                       & 64.10          & 67.35          & \textbf{69.07}        & 79.02       & \textbf{86.72}    & 93.96           & \textbf{83.37}    & 77.65                     \\ \midrule
                                                     & Qwen3-VL                & 2B                      & 70.29          & 50.42          & 49.35        & 64.69       & 78.09    & 91.45           & 80.43    & 69.24                     \\
                                                     & Qwen2.5-VL              & 3B                      & 63.03          & 47.47          & 54.87        & 66.66       & 80.57    & 86.85           & 82.84    & 68.90                     \\
                                                     & AnomalyR1               & 3B                      & 60.62          & 63.56          & \textbf{70.14}        & \textbf{80.47}       & 85.28    & 92.48           & 86.15    & 76.96                     \\
                                                     & Qwen3-VL                & 4B                      & 74.02          & 59.25          & 62.74        & 70.09       & 79.22    & 92.47           & 83.59    & 74.48                     \\
                                                     & AnomalyGPT              & 7B                      & 65.57          & 27.49          & 27.97        & 36.86       & 32.11    & 29.84           & 35.82    & 36.52                     \\
                                                     & Qwen2.5-VL              & 7B                      & 71.79          & 55.23          & 68.97        & 64.54       & 78.90    & 92.55           & 84.66    & 73.81                     \\
                                                     & LLaVA-1.5               & 7B                      & 51.33          & 37.04          & 36.62        & 50.60       & 69.79    & 68.29           & 69.53    & 54.74                     \\
                                                     & LLaVA-NEXT              & 7B                      & 57.64          & 33.79          & 47.72        & 51.84       & 67.93    & 81.39           & 74.91    & 59.32                     \\
                                                     & LLaVA-OneVision         & 7B                      & 51.77          & 46.13          & 41.85        & 62.19       & 69.73    & 90.31           & 80.93    & 63.27                     \\
                                                     & InternVL2               & 8B                      & 59.97          & 43.85          & 47.91        & 57.60       & 78.10    & 74.18           & 80.37    & 63.14                     \\
                                                     & Qwen3-VL                & 8B                      & 73.08          & 62.33          & 60.34        & 69.15       & 80.09    & 92.53           & 84.01    & 74.51                     \\
                                                     & Qwen3-VL-thinking       & 8B                      & 52.31          & 41.50          & 40.37        & 45.67       & 36.02    & 64.34           & 53.30    & 47.64                     \\
                                                     & GLM-4.1V-Thinking       & 9B                      & 72.31          & 72.01          & 66.47        & 79.98       & 84.61    & 93.79           & 83.26    & 78.92                     \\
                                                     & LLaVA-1.5               & 13B                     & 49.96          & 38.78          & 46.17        & 58.17       & 73.09    & 73.62           & 70.98    & 58.68                     \\
                                                     & Kimi-VL               & 16B                     & 66.57          & 71.72          & 61.88        & 79.78       & 82.93    & 92.18           & \textbf{86.51}    & 77.37                     \\
                                                     & LLaVA-NeXT              & 34B                     & 57.92          & 48.79          & 52.87        & 71.34       & 80.28    & 81.12           & 77.80    & 67.16                     \\
\multirow{-16}{*}{Open Source}                       & InternVL2               & 76B                     & 68.25          & 54.22          & 56.66        & 66.30       & 80.47    & 86.40           & 82.92    & 70.75                     \\ \hline
                                                     & Qwen3-VL                & 2B                      & 61.90          & 70.20          & 54.63        & 71.51       & 79.92    & 94.59           & 81.61    & 73.48                     \\
                                                     & Qwen3-VL                & 4B                      & 75.05          & 73.72          & 66.48        & 76.83       & 82.33    & 94.90           & 84.41    & 79.10                     \\
                                                     & Qwen2.5-VL              & 7B                      & 70.74          & 67.97          & 60.82        & 72.78       & 82.62    & \textbf{97.17}           & 84.99    & 76.73                     \\
\multirow{-4}{*}{Reason-IAD}                         & Qwen3-VL                & 8B                      & \textbf{75.24}          & \textbf{75.07}          & 62.78        & 76.38       & \textbf{85.58}    & 96.31           & 84.62    & \textbf{79.43}                     \\ \bottomrule
\end{tabular}
}
\end{table*}

\textbf{Dynamic Visual Injection.}
Inspired by human cognition, in which uncertain situations often prompt revisiting visual information, we introduce a dynamic visual injection mechanism within the latent reasoning process. This mechanism selectively incorporates the most informative image patches into the latent sequence alongside the latent think tokens, guided by the reward signal at each iteration.

Specifically, we leverage the attention weights of the latent think tokens to identify $t$ highly relevant image patches. At each iteration, the model resamples $t$ candidate patches $V_{cand} = \{v_1, \dots, v_t\}$ according to the attention distribution computed from the current latent sequence, and evaluates their contribution by injecting them into the latent sequence.
The optimal patch set $V_{best}$ is then updated based on the current reward $\mathcal{R}$ as follows:
\begin{equation}
V_{best} = 
\begin{cases} 
V_{cand}, & \text{if } \mathcal{R} > \mathcal{R}_{best}, \\
V_{best}, & \text{otherwise}.
\end{cases}
\end{equation}

If the current reward $\mathcal{R}$ exceeds $\mathcal{R}_{best}$, the candidate patches are considered to provide more informative visual evidence (e.g., clearer defect patterns or more discriminative local details), and the best patch set $V_{best}$ is updated accordingly. Otherwise, the previous best patches are retained.
As shown in Algorithm~\ref{alg:reason_iad} in the Appendix, through this iterative, reward-guided refinement, the selected visual patches progressively concentrate on regions most relevant to the latent think tokens, thereby guiding the latent reasoning process toward more accurate and efficient anomaly detection.

\section{Experiments}
\subsection{Experiment Setup}
\textbf{Compared Baselines.} To comprehensively evaluate the reasoning capabilities of the proposed Reason-IAD, we conduct extensive experiments comparing it against a diverse set of representative models, including both commercial and open-source multimodal large language models (MLLMs) as well as anomaly detection–specific models. The commercial baselines include Claude-3.5-Sonnet~\cite{Anthropic2024Claude3}, Gemini-2.5-Pro~\cite{comanici2025gemini}, GPT-4o~\cite{hurst2024gpt}, and GPT-5-Mini~\cite{singh2025openai}. The open-source baselines include Qwen2.5-VL~\cite{bai2025qwen2}, Qwen3-VL~\cite{bai2025qwen3vltechnicalreport}, LLaVA-1.5~\cite{liu2024improved}, LLaVA-NeXT~\cite{liu2024llava}, LLaVA-OneVision~\cite{li2024llava}, and InternVL2~\cite{chen2024internvl}. In addition, we evaluate Reason-IAD against anomaly detection–specific models, including AnomalyGPT~\cite{gu2024anomalygpt} and Anomaly-R1~\cite{chao2025anomalyr1}.

\begin{table*}[]
\caption{Performance comparison of open-source MLLMs on the MMAD benchmark under the standard zero-shot setting. The best results are highlighted in bold, and all metrics are reported as accuracy (\%).}
\label{table:0shot}
\resizebox{\linewidth}{!}{
\begin{tabular}{ccccccccccc}
\toprule
\multirow{2}{*}{Type}         & \multirow{2}{*}{Model} & \multirow{2}{*}{Scale} & Anomaly              & \multicolumn{4}{c}{Defect}                                                                & \multicolumn{2}{c}{Object}                  & \multirow{2}{*}{Average} \\ \cline{4-10}
                              &                        &                        & Discrimination       & Classification       & Localization         & Description          & Analysis             & Classification       & Analysis             &                          \\ \hline
\multirow{12}{*}{Open Source} & Qwen3-VL               & 2B                     & 63.27                & 47.71                & 52.48                & 63.48                & 77.82                & 92.16                & 80.03                & 68.14                    \\
                              & Qwen2.5-VL             & 3B                     & 60.63                & 47.03                & 54.44                & 63.51                & 79.25                & 87.48                & 82.02                & 67.77                    \\
                              & Qwen3-VL               & 4B                     & 65.83                & 53.97                & 58.16                & 66.35                & 77.82                & 91.98                & 83.38                & 71.07                    \\
                              & Qwen3-VL thinking      & 4B                     & 64.21                & 39.61                & 42.98                & 48.96                & 38.32                & 78.18                & 64.18                & 53.78                    \\
                              & Qwen2.5-VL             & 7B                     & 60.35                & 49.89                & 57.51                & 60.41                & 75.58                & 93.39                & 84.80                & 68.85                    \\
                              & LLaVA-1.5              & 7B                     & 50.81                & 36.82                & 35.94                & 50.96                & 70.45                & 67.33                & 69.12                & 54.49                    \\
                              & InternVL3              & 8B                     & 64.67                & 50.91                & 60.06                & 65.49                & 77.16                & 82.01                & 83.58                & 69.13                    \\
                              & Qwen3-VL               & 8B                     & 67.10                & 57.98                & 58.86                & 66.89                & 77.65                & 92.94                & 83.82                & 72.18                    \\
                              & Qwen3-VL thinking      & 8B                     & 63.07                & 45.81                & 49.56                & 53.52                & 40.63                & 82.51                & 66.25                & 57.34                    \\
                              & LLaVA-NeXT             & 34B                    & 60.31                & 51.40                & 55.47                & 71.62                & 80.49                & 81.45                & 75.08                & 67.97                    \\
                              & Qwen2.5-VL             & 72B                    & 66.22                & 57.99                & \textbf{62.95}                & 72.21                & 81.01                & 93.92                & \textbf{86.51}                & 74.40                    \\
                              & InternVL2              & 76B                    & 64.30                & 51.19                & 54.20                & 63.46                & 79.92                & 89.34                & 83.48                & 69.41                    \\ \hline
\multirow{4}{*}{Reason-IAD}   & Qwen3-VL               & 2B                     & 54.66                & 68.95                & 55.28                & 70.64                & 79.68                & 95.28                & 80.85                & 72.19                    \\
                              & Qwen3-VL               & 4B                     & 66.58                & 70.62                & 62.82                & 73.28                & \textbf{80.88}                & 93.80                & 84.88                & 76.12                    \\
                              & Qwen2.5-VL             & 7B                     & 59.23                & 61.02                & 55.39                & 66.57                & 78.37                & \textbf{97.36}                & 85.50                & 71.92                    \\
                              & Qwen3-VL               & 8B                     & \textbf{67.73}                & \textbf{73.12}                & 59.74                & \textbf{74.69}                & 80.65                & 96.78                & 84.24                & \textbf{76.71}                    \\ \bottomrule   
\end{tabular}
}
\end{table*}

\textbf{Benchmarks.} We evaluate Reason-IAD on the MMAD benchmark~\cite{jiang2024mmad}, which assesses performance across seven subtasks: anomaly discrimination, defect classification, defect localization, defect description, defect analysis, object classification, and object analysis. MMAD comprises four datasets: MVTec-AD~\cite{bergmann2019mvtec}, VisA~\cite{zou2022spot}, MVTec-LOCO~\cite{bergmann2022beyond}, and GoodsAD~\cite{zhang2024pku}. 

\textbf{Implementation Details.} In this study, we adopt Qwen2.5-VL-7B and Qwen3-VL-2B/4B/8B as the base models. For retrieval, we employ CLIP with a ViT-L/14-336 backbone as the image and text encoders, with top-$k$ set to 2. The number of latent thinking tokens $\mathcal{Z}$ is set to 4, and the default number of reasoning iterations is 10. The learning rate is fixed at $1 \times 10^{-3}$. To ensure stable exploration in the latent space, we set the perturbation magnitude to 10\%. By default, we employ a one-shot setting with a randomly sampled normal image from the same domain as the reference template. For comparison, we also evaluate a zero-shot setting. All experiments are conducted on four NVIDIA A100 GPUs (80GB). More details are provided in the Appendix.


\subsection{Main Results}

Table~\ref{table:1shot} presents a comprehensive comparison of Reason-IAD with state-of-the-art methods under the one-shot setting. Overall, Reason-IAD demonstrates strong performance across all seven tasks. On Qwen3-VL-8B, it achieves an average accuracy of 79.43\%, improving 4.92\% over the base model, and even surpasses the commercial model GPT-4o (74.92\%). Although AnomalyGPT is designed for anomaly detection, it underperforms on classification and localization, which may be attributed to catastrophic forgetting when trained on expert outputs. Another domain-specific method, AnomalyR1, incorporates GRPO to enhance anomaly understanding, yet our training-free method still outperforms it by 2.47\% in average accuracy, demonstrating stronger reasoning capability. Notably, Reason-IAD surpasses standard human annotators on defect analysis and object classification tasks. Furthermore, while Qwen3-VL-thinking relies on explicit reasoning chains that generate lengthy intermediate outputs, our implicit reasoning achieves a 31.79\% gain in performance, highlighting the effectiveness and efficiency of Reason-IAD.

Table~\ref{table:0shot} reports the results under the zero-shot setting, where no normal reference images are provided. Reason-IAD consistently improves the average performance of the evaluated base models, with Qwen3-VL-4B and Qwen3-VL-8B achieving average accuracy gains of 5.05\% and 4.53\%, respectively. Notably, our 8B model outperforms InternVL2-76B by 7.30\%, further demonstrating the effectiveness of Reason-IAD in anomaly understanding.

\subsection{Model Analysis}
\begin{figure}[t]
    \centering
    \begin{minipage}{0.35\linewidth}
        \centering
        \includegraphics[width=\linewidth]{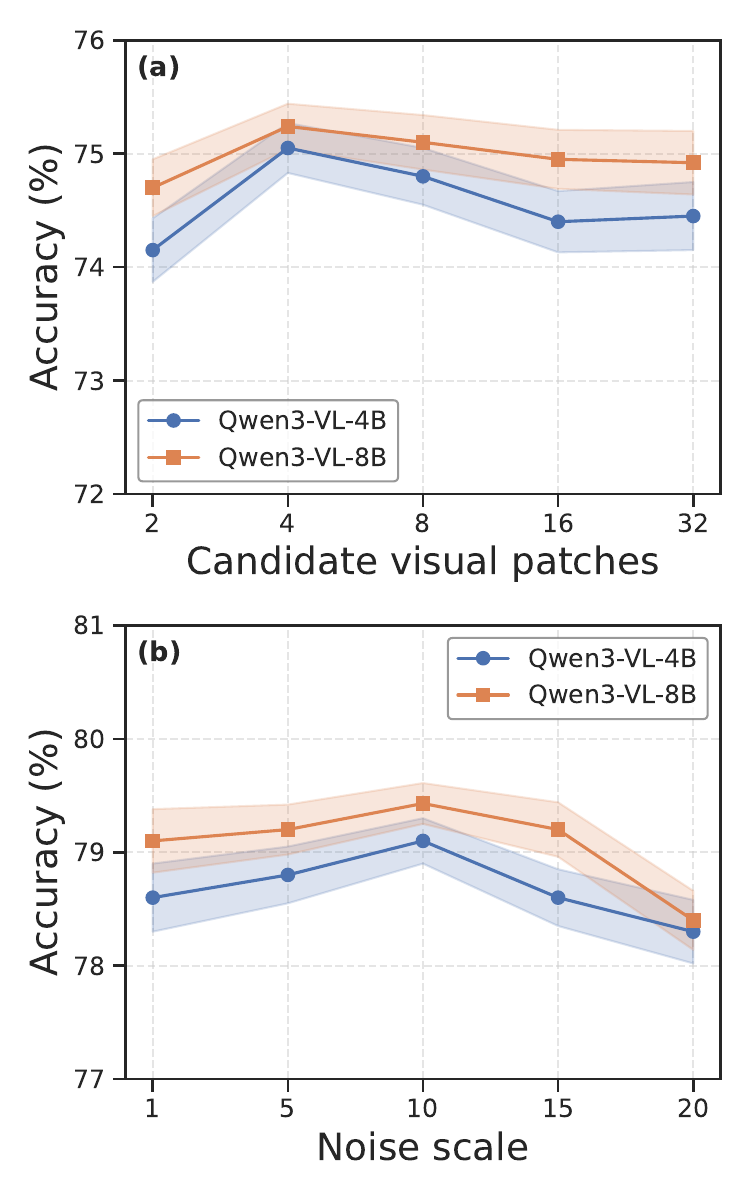}
        \caption{(a) Effect of the number of injected candidate visual patches. (b) Impact of noise magnitude.}
        \label{fig:patch}
    \end{minipage}
    \hfill
    \begin{minipage}{0.6\linewidth}
        \centering
        \includegraphics[width=\linewidth]{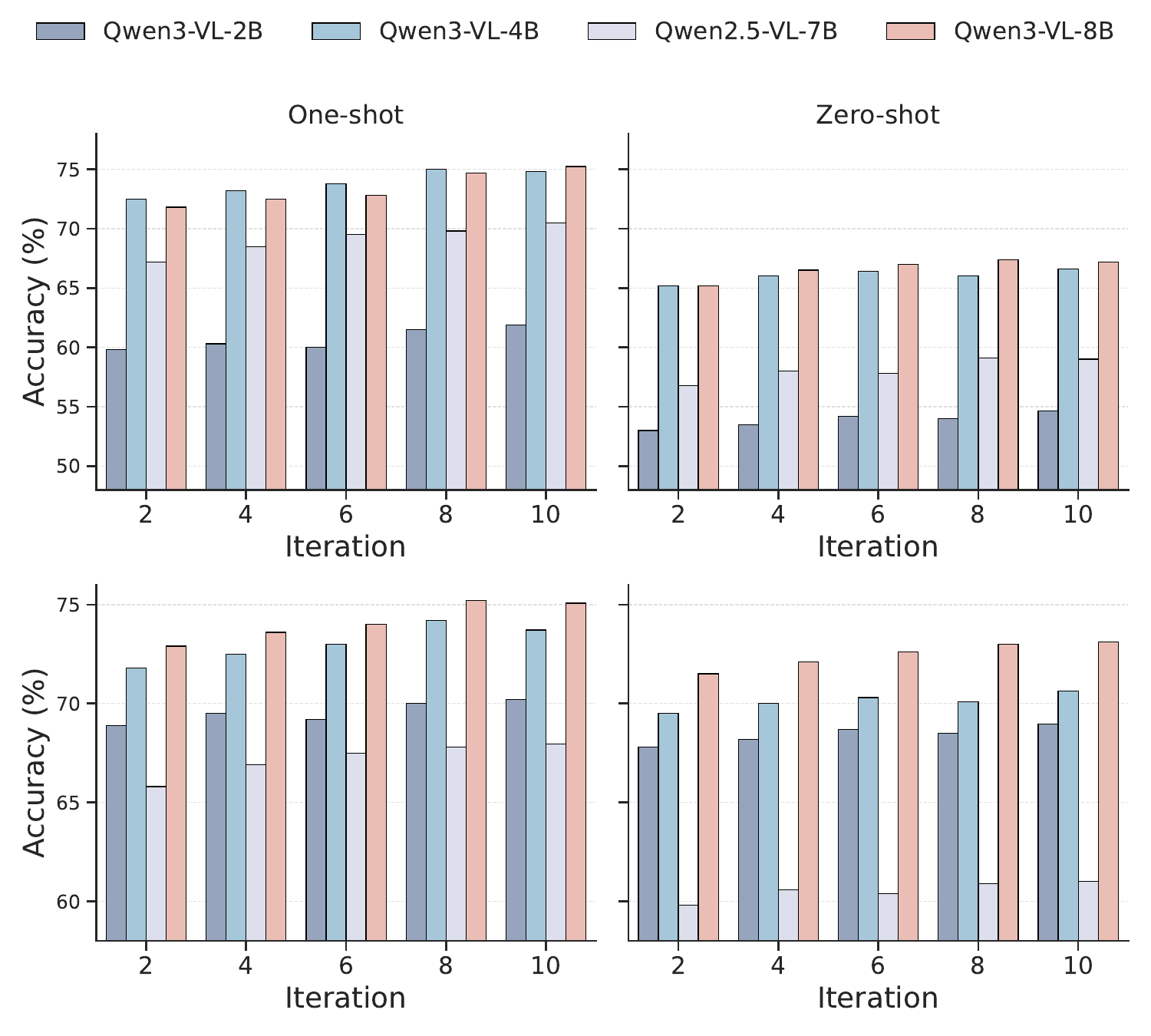}
        \caption{Impact of iteration count on anomaly discrimination (top row) and anomaly classification performance (bottom row) in one-shot and zero-shot settings.}
        \label{fig:iteration}
    \end{minipage}
\end{figure}

\textbf{Ablation Study.} To evaluate the contributions of key components in the proposed Reason-IAD, we conduct ablation studies on four public datasets, as summarized in Table~\ref{table:ablation}. Incorporating knowledge guidance consistently improves detection accuracy across all datasets. Specifically, on MVTec, accuracy increases from 85.75\% to 88.49\%, with an average gain of 3.22\% across the four datasets, highlighting the critical role of prior knowledge. Latent reasoning further enhances performance by 1.54\%, improving the model’s inference capability. Moreover, integrating visual information provides an additional average improvement of 0.32\%, indicating that the model effectively leverages visual evidence during reasoning. When all components are combined, Reason-IAD achieves the best overall performance, underscoring the effectiveness of the proposed modules.


\begin{table}[h]
\caption{Ablation study of different components with Qwen3-VL-8B under the one-shot setting on four public anomaly detection datasets, all metrics are reported as accuracy (\%).}
\label{table:ablation}
\centering
\begin{tabular}{ccccccc}
\toprule
\multicolumn{3}{c}{Module} & \multicolumn{4}{c}{Datasets} \\ \hline
RAKI & EDLR & $V_{patch}$ & MVTec & VisA & MVTec-LOCO & GoodsAD \\ \hline
  \textcolor{black!40}{\ding{55}}  &  \textcolor{black!40}{\ding{55}}   &  \textcolor{black!40}{\ding{55}}   & 85.75 & 75.44 & 67.45 & 69.38 \\
\ding{51} &  \textcolor{black!40}{\ding{55}}   &  \textcolor{black!40}{\ding{55}}   & 88.49 & 77.83 & 70.76 & 73.83 \\
\ding{51} & \ding{51} &  \textcolor{black!40}{\ding{55}}   & 90.03 & 79.13 & 71.95 & 75.29 \\
\rowcolor[HTML]{DAE8FC}
\ding{51} & \ding{51} & \ding{51} & \textbf{90.36} & \textbf{79.46} & \textbf{72.29} & \textbf{75.61} \\ \bottomrule
\end{tabular}
\end{table}

\textbf{Impact of Visual Patch Number.} As shown in Figure~\ref{fig:patch}(a), we study the effect of varying the number of injected visual patches on anomaly detection performance. As the number of patches increases, detection accuracy improves steadily; however, further increasing the number leads to performance degradation. This suggests that a limited set of fine-grained visual cues is sufficient to effectively guide latent reasoning, whereas excessive patches introduce redundancy that hinders optimization.

\textbf{Impact of Noise Scale.} We further investigate how the perturbation magnitude $\sigma$ governs the dynamics of latent optimization. As illustrated in Figure~\ref{fig:patch}(b), increasing the initial noise scale enhances exploration, enabling the model to traverse a broader range of latent trajectories and discover higher-confidence reasoning paths. However, when $\sigma$ becomes excessively large, the perturbations destabilize the optimization process, resulting in performance degradation. This suggests that latent reasoning benefits from a well-calibrated level of perturbation.

\textbf{Impact of Iteration Number.} To evaluate the effect of iteration count on model performance, we conduct experiments on four base models for anomaly detection, as shown in Figure~\ref{fig:iteration}. The results indicate that increasing the number of iterations consistently improves anomaly reasoning, demonstrating that iterative optimization effectively enhances latent reasoning. Although minor fluctuations occur in intermediate steps, the models maintain high accuracy and continue to benefit from additional iterations, highlighting their capacity to leverage iterative optimization. Iterations are fixed at 10 to balance performance and efficiency.
\vspace{-1em}
\begin{wraptable}{r}{0.6\textwidth}
\centering
\caption{Ablation study on the number of latent tokens.}
\label{tab:latent_tokens}
\setlength{\tabcolsep}{5.5pt}
\begin{tabular}{cccc}
\hline
\makecell{Latent \\ Tokens} 
& \makecell{Anomaly \\ Discrimination} 
& \makecell{Defect \\ Localization} 
& \makecell{Object \\ Classification} \\
\hline
2 & 73.91 & 61.12 & 95.64 \\
3 & 74.68 & 62.01 & 96.02 \\
\rowcolor[HTML]{DAE8FC}
4 & \textbf{75.24} & \textbf{62.78} & \textbf{96.31} \\
5 & 74.97 & 62.43 & 96.18 \\
6 & 74.63 & 62.05 & 95.91 \\
\hline
\end{tabular}
\end{wraptable}

\textbf{Sensitivity to the Number of Latent Tokens.} To evaluate how the number of latent tokens influences the capacity of the latent reasoning space, we perform an ablation study by varying the token count, with results shown in Table~\ref{tab:latent_tokens}. An insufficient number of tokens limits the representation of intermediate reasoning states, whereas an excessive number enlarges the optimization space and slightly compromises stability. Performance remains relatively stable across configurations, with four latent tokens yielding the best results. This indicates that a moderate number of latent tokens strikes a favorable balance between reasoning capacity and optimization stability.


\begin{figure}
    \centering
    \includegraphics[width=\linewidth]{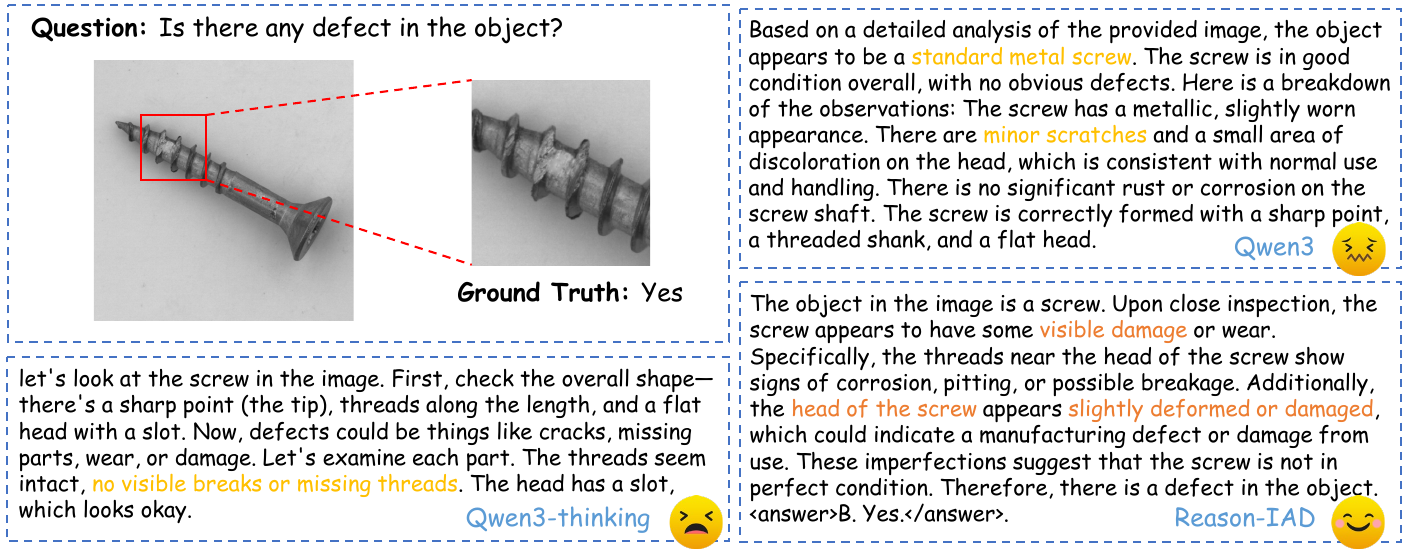}
    \caption{Comparison of model outputs for anomaly detection.}
    \label{fig:case}
\end{figure}


\textbf{Qualitative Analysis.}
Figure~\ref{fig:case} presents a qualitative comparison of different MLLMs on an industrial anomaly detection case. Both Qwen3-VL and Qwen3-VL-thinking fail to correctly identify the defective screw. Specifically, Qwen3-VL-thinking performs step-by-step reasoning but overlooks subtle defects in the screw structure, while Qwen3-VL interprets visible scratches and discoloration as normal wear, leading to an incorrect prediction. In contrast, Reason-IAD correctly identifies the anomaly by recognizing potential damage around the screw head and threads, providing a more accurate and detailed explanation. These results suggest that explicit reasoning alone may be insufficient for fine-grained anomaly recognition, whereas integrating domain-specific knowledge and relevant visual evidence can enhance anomaly understanding and reasoning reliability.

\begin{table}[t]
\centering

\begin{minipage}[t]{0.45\linewidth}
\centering
\caption{Comparison of retrieval strategies.}
\label{tab:raki_strategy}
\vspace{0.2mm}

\setlength{\tabcolsep}{6pt}
\begin{tabular}{@{}cccc@{}}
\toprule
Strategy & MVTec & VisA & Avg. \\
\midrule
w/o RAKI & 87.03 & 76.88 & 81.96 \\
Top-1 & 89.91 & 78.74 & 84.33 \\
\rowcolor[HTML]{DAE8FC}
Top-2 & \textbf{90.36} & \textbf{79.46} & \textbf{84.91} \\
Top-3 & 89.72 & 78.96 & 84.34 \\
Random-2 & 86.41 & 75.73 & 81.07 \\
\bottomrule
\end{tabular}

\end{minipage}
\hfill
\begin{minipage}[t]{0.54\linewidth}
\centering
\caption{Comparison of reward formulations for latent reasoning optimization.}
\label{tab:reward_strategy}
\setlength{\tabcolsep}{6pt}
\begin{tabular}{@{}cccc@{}}
\toprule
Reward & MVTec & VisA & Avg. \\
\midrule
w/o latent optimization & 88.49 & 77.83 & 83.16 \\
Random reward & 88.12 & 77.46 & 82.79 \\
Max-probability reward & 89.71 & 78.83 & 84.27 \\
\rowcolor[HTML]{DAE8FC}
Entropy-based reward & \textbf{90.36} & \textbf{79.46} & \textbf{84.91} \\
\bottomrule
\end{tabular}
\end{minipage}

\end{table}

\textbf{Sensitivity to Knowledge Retrieval Strategies.}
We evaluate the impact of different retrieval strategies on anomaly reasoning performance. As shown in Table~\ref{tab:raki_strategy}, Top-2 achieves the highest average accuracy of 84.91\%, outperforming Top-1 by 0.58 percentage points. This suggests that an additional candidate can improve robustness to category-ranking errors. However, increasing the number of retrieved categories to three slightly degrades performance, indicating potential interference from irrelevant knowledge. Moreover, Random-2 underperforms even the setting without RAKI, highlighting the importance of retrieval relevance.

\textbf{Comparison of Reward Strategies.}
We investigate the effectiveness of different reward formulations for latent reasoning. As shown in Table~\ref{tab:reward_strategy}, random rewards fail to improve performance, whereas both maximum-probability and entropy-based rewards yield consistent gains. In particular, the entropy-based reward achieves the highest average accuracy of 84.91\%, surpassing the maximum-probability reward by 0.64 percentage points. Unlike maximum-probability rewards that focus solely on the most likely prediction, entropy-based rewards capture uncertainty across the entire candidate-answer distribution, providing a more comprehensive signal for latent exploration.

\section{Conclusion}
In this work, we introduce Reason-IAD, a knowledge-guided dynamic latent reasoning framework for explainable industrial anomaly detection. By integrating a retrieval-augmented knowledge module, an entropy-driven latent reasoning mechanism, and a dynamic visual injection strategy, Reason-IAD facilitates context-aware, iterative reasoning over fine-grained industrial defects while effectively leveraging critical visual evidence. Extensive experiments demonstrate that Reason-IAD consistently outperforms state-of-the-art methods, achieving high detection accuracy and interpretable reasoning. In summary, Reason-IAD provides a promising paradigm for training-free, expert-level anomaly detection. Future directions include adaptive knowledge retrieval and advanced latent optimization strategies to further enhance the applicability of MLLMs across diverse industrial scenarios.


\begin{ack}
\textbf{Funding.}
This work was supported in part by the National Natural Science Foundation of China (No. 62301621, No. 62672529, No. 62625602, 62411540034), Shenzhen Science and Technology Program (No. 20231121172359002, 2023A008, NO.SYSRD20250529113401002), Shenzhen General Research Project (No. JCYJ20241202125904007), Guangdong Basic and Applied Basic Research Foundation (No. 2025A1515011398), and Guangdong Major Project of Basic Research (2023B0303000010).

\textbf{Competing Interests.}
The authors declare that they have no competing interests related to this work.
\end{ack}



{\small
\bibliographystyle{unsrt}
\bibliography{references}
}







\newpage
\appendix

\section*{Appendix}
This appendix provides supplementary material for the main paper. We present detailed experimental settings, datasets, prompt design, and additional experimental results, followed by further case study analyses. Finally, we discuss the potential impact and limitations of the proposed method.

The supplementary material is organized as follows:
\begin{itemize}
    \item Sec.~\ref{More Details} provides additional details about the evaluation framework, including experimental setup, datasets, prompt design, and an overview of the proposed method.
    \item Sec.~\ref{Additional Experimental Results} presents additional experimental results.
    \item Sec.~\ref{Case Study} provides case studies to further illustrate the effectiveness of the proposed method.
    \item Sec.~\ref{Impact and Limitation} discusses the broader impact and limitations of the proposed approach.
\end{itemize}

\section{More Details about Evaluation}
\label{More Details}
\subsection{Experiment Setup.}
To comprehensively evaluate the reasoning capabilities of the proposed Reason-IAD, we conduct extensive experiments comparing it against a diverse set of representative models. These include both commercial and open-source multimodal large language models (MLLMs), as well as models specifically designed for anomaly detection. The commercial baselines include Claude-3.5-Sonnet~\cite{Anthropic2024Claude3}, Gemini-2.5-Pro~\cite{comanici2025gemini}, GPT-4o~\cite{hurst2024gpt}, and GPT-5-Mini~\cite{singh2025openai}. The open-source baselines include Qwen2.5-VL~\cite{bai2025qwen2}, Qwen3-VL~\cite{bai2025qwen3vltechnicalreport}, LLaVA-1.5~\cite{liu2024improved}, LLaVA-NeXT~\cite{liu2024llava}, LLaVA-OneVision~\cite{li2024llava}, and InternVL2~\cite{chen2024internvl}. In addition, we benchmark Reason-IAD against anomaly detection–specific models, such as AnomalyGPT~\cite{gu2024anomalygpt} and Anomaly-R1~\cite{chao2025anomalyr1}, to evaluate its performance relative to domain-specialized approaches.

For our experiments, we use Qwen2.5-VL-7B and Qwen3-VL-2B/4B/8B as base models. In the retrieval stage, we employ the CLIP model with a ViT-L/14-336 backbone as the image and text encoder, with top-$k$ set to 2. The number of latent thinking tokens $\mathcal{Z}$ is set to 4, and the default number of reasoning iterations is 10. The learning rate is fixed at $1 \times 10^{-3}$, and the perturbation magnitude is set to 10\% to ensure stable exploration in the latent space. By default, we adopt a one-shot setting, where a normal image from the same domain is randomly sampled as a reference template to help the model capture the normal appearance of objects. For comparison, we also evaluate the models under a zero-shot setting, which tests their ability to perform pairwise comparisons between industrial images without reference guidance. This evaluation protocol provides a comprehensive assessment of the models’ reasoning capabilities in both reference-based and reference-free scenarios. All experiments are conducted on four NVIDIA A100 GPUs (80GB).

\subsection{Datasets.}

\begin{table*}[h]
\centering
\caption{Statistics on the composition and quantity of test data.}
\label{table:datasets}
\resizebox{\linewidth}{!}{
\renewcommand{\arraystretch}{1.2}
\begin{tabular}{ccccc}
\hline
Image Source  & Sampled Images & Generated Questions & Object Categories & Defect Categories \\ \hline
MVTec-AD~\cite{bergmann2019mvtec}      & 1691           & 8338                & 15                & 73                \\
MVTec-LOCO-AD~\cite{bergmann2022beyond} & 1566           & 7624                & 5                 & 89                \\
VisA~\cite{zou2022spot}          & 2141           & 10622               & 12                & 67                \\
GoodsAD~\cite{zhang2024pku}       & 2968           & 13088               & 6                 & 15                \\ \hline
\textbf{SUM}           & 8366           & 39672               & 38                & 244               \\ \hline
\end{tabular}
}
\end{table*}

We evaluate Reason-IAD on multiple industrial anomaly detection (IAD) datasets with diverse characteristics, covering over 38 product categories and 244 defect types, as summarized in Table~\ref{table:datasets}. Among these, MVTec AD is one of the most widely used benchmarks for IAD, encompassing a variety of object and texture categories; we further leverage fine-grained annotations from Defect Spectrum to provide more detailed defect descriptions. MVTec LOCO AD focuses on logical anomalies, thereby evaluating the model’s ability to reason about high-level logical inconsistencies. The VisA dataset contains multiple instances with complex defect patterns, reflecting more challenging real-world IAD scenarios. GoodsAD primarily consists of industrial goods, and due to the diverse appearances of finished products across different brands, it significantly increases the number of object categories and visual variations.

In our experiments, we adopt the MMAD benchmark as the evaluation protocol, which comprises seven subtasks:
(1) Anomaly Discrimination Detection, a binary classification task that determines whether a sample is defective, to evaluate the model’s fundamental anomaly detection capability; 
(2) Defect Classification, which identifies the specific defect type, jointly assessing anomaly recognition and domain knowledge of industrial defect categories; 
(3) Defect Localization, requiring the model to locate defective regions, where standardized textual descriptions are used instead of explicit segmentation masks; 
(4) Defect Description, which characterizes defect attributes (e.g., size and color), simulating practical defect inspection scenarios; 
(5) Defect Analysis, analyzing the potential impact of defects on product quality to estimate their severity; 
(6) Object Classification, categorizing industrial products to facilitate anomaly detection based on an understanding of normal object characteristics; and 
(7) Object Analysis, querying the composition, position, and function of products to evaluate fine-grained semantic understanding.

\begin{algorithm}[t]
\caption{Reason-IAD: Knowledge-Guided Dynamic Latent Reasoning}
\label{alg:reason_iad}
\begin{algorithmic}[1]
\STATE {\bfseries Input:} Query image $I$, question $Q$, knowledge repository 
$\mathcal{K}=\{(c_i,\mathrm{desc}_i)\}_{i=1}^N$, 
latent token number $m$, iteration number $N_{\text{iter}}$, top-$k$ categories

\STATE Encode image: $\mathbf{v} \gets f_v(I)$
\FOR{each category $c_i \in \mathcal{K}$}
    \STATE Encode text: $\mathbf{k}_i \gets f_t(c_i)$
    \STATE Compute similarity: $s_i \gets \cos(\mathbf{v}, \mathbf{k}_i)$
\ENDFOR
\STATE Retrieve top-$k$ categories $\{c_{(1)},\dots,c_{(k)}\}$ by $\{s_i\}$ and Construct enriched prompt
\STATE Initialize latent think tokens $\mathcal{Z}=\{z_1,\dots,z_m\}$
\FOR{$n=1$ to $N_{\text{iter}}$}
    \STATE Sample noise $\xi^{(n)} \sim \mathcal{N}(0,\sigma^2 I)$
    \STATE $\mathcal{Z}'^{(n)} \leftarrow \mathcal{Z}^{(n)} + \eta \nabla_{\mathcal{Z}^{(n)}} J(\mathcal{Z}^{(n)})$
    \STATE Initialize best patch $V_{best}$
    \FOR{$j=1 ,\dots,m$}
    \STATE Sample candidate patches $V_{\text{cand}}$ according to attention distribution
    \STATE Compute entropy $\mathcal{H}^{(n)}$ and reward $\mathcal{R}^{(n)}$ using Eq.~\eqref{eq:entropy}--\eqref{eq:reword}
    \IF{$\mathcal{R}^{(n)} > \mathcal{R}_{\text{best}}$}
        \STATE $V_{\text{best}} \gets V_{\text{cand}}$
        \STATE $\mathcal{R}_{\text{best}} \gets \mathcal{R}^{(n)}$
    \ENDIF
    \ENDFOR
\ENDFOR
\STATE \textbf{Return} reasoning result
\end{algorithmic}
\end{algorithm}

\subsection{Overview of the Proposed Method}

In this work, we introduce Reason-IAD, a knowledge-guided dynamic latent reasoning framework for explainable industrial anomaly detection, as illustrated in Algorithm~\ref{alg:reason_iad}. By integrating a retrieval-augmented knowledge module, an entropy-driven latent reasoning mechanism, and a dynamic visual injection strategy, Reason-IAD enables context-aware iterative reasoning over fine-grained industrial defects while effectively leveraging critical visual evidence.

Extensive experiments demonstrate that Reason-IAD consistently outperforms state-of-the-art methods, achieving high detection accuracy and improved interpretability. Overall, Reason-IAD provides a promising paradigm for training-free, expert-level industrial anomaly detection.

\subsection{Prompt Design}
To guide the multimodal large language models (MLLMs) to perform domain-specific anomaly reasoning, we design a structured prompt that explicitly defines the model as an industrial inspection expert and incorporates category-level domain knowledge. Specifically, the prompt instructs the model to determine whether the queried image contains defects and to answer the associated questions based on visual evidence. To encourage reliable decision-making, the model is required to perform internal reasoning before producing the final prediction.

Moreover, we inject retrieved domain knowledge describing typical defect types and normal object characteristics into the prompt, enabling context-aware reasoning tailored to industrial scenarios. This design allows the model to leverage both visual observations and prior knowledge, facilitating accurate and interpretable anomaly detection. The prompt template is defined as follows:
\begin{quote}
You are an expert industrial inspector responsible for analyzing product images. Your task is to determine whether the query image contains any defects and to answer the related questions. You should first perform the reasoning process internally and then provide the final answer. The following domain knowledge describes typical defect types and normal object characteristics: \verb|\n| \{\texttt{Domain\_knowledge}\}.
\end{quote}

\section{Additional Experimental Results}
\label{Additional Experimental Results}

\begin{table*}[]
\centering
\caption{Performance comparison of different MLLMs in Anomaly Discrimination Tasks.}
\label{table:f1}
\begin{tabular}{cccccc}
\toprule
                              & Model                  & Scale      & Accuracy & Recall & Precision \\ \hline
\multirow{2}{*}{Human}        & Human (expert)         & \textbf{-} & 95.24    & 94.25  & 98.89     \\
                              & Human (ordinary)       & \textbf{-} & 86.90    & 87.07  & 94.35     \\ \hline
\multirow{5}{*}{Commercial}   & claude-3.5-sonnet      & -          & 60.14    & 30.87  & 76.75     \\
                              & Gemini-1.5-flash       & -          & 58.58    & 78.63  & 67.41     \\
                              & Gemini-1.5-pro         & -          & 68.63    & 45.47  & 86.84     \\
                              & GPT-4o-mini            & -          & 64.33    & 65.47  & 73.04     \\
                              & GPT-4o                 & -          & 68.63    & 67.37  & 75.68     \\ \hline
\multirow{17}{*}{Open Source} & Qwen3-VL               & 2B         & 70.29    & 66.83  & 79.78     \\
                              & Qwen3-VL               & 4B         & 74.02    & 57.95  & 89.41     \\
                              & AnomalyGPT             & 7B         & 65.57    & 82.11  & 74.45     \\
                              & Qwen2.5-VL             & 7B         & 71.79    & 53.74  & 75.54     \\
                              & Qwen-VL-Chat           & 7B         & 53.65    & 43.95  & 65.39     \\
                              & LLaVA-1.5              & 7B         & 51.33    & 94.79  & 62.72     \\
                              & Qwen3-VL               & 8B         & 73.08    & 75.58  & 79.76     \\
                              & Cambrian-1*            & 8B         & 55.60    & 22.28  & 74.10     \\
                              & SPHINX*                & 7B         & 53.13    & 6.42   & 99.74     \\
                              & LLaVA-NEXT-Interleave  & 7B         & 57.64    & 16.58  & 90.83     \\
                              & InternLM-XComposer2-VL & 7B         & 55.85    & 17.94  & 75.87     \\
                              & LLaVA-OneVision         & 7B         & 51.77    & 4.90   & 78.19     \\
                              & MiniCPM-V2.6           & 8B         & 57.31    & 34.38  & 70.98     \\
                              & InternVL2              & 8B         & 59.97    & 30.25  & 79.22     \\
                              & LLaVA-1.5              & 13B        & 49.96    & 99.79  & 62.00     \\
                              & LLaVA-NeXT             & 34B        & 57.92    & 46.27  & 69.98     \\
                              & InternVL2              & 76B        & 68.25    & 55.81  & 83.52     \\ \hline
\multirow{4}{*}{Reason-IAD}   & Qwen3-VL               & 2B         & 61.90    & 92.23  & 69.19     \\
                              & Qwen3-VL               & 4B         & 75.05    & 68.21  & 86.40     \\
                              & Qwen2.5-VL             & 7B         & 70.74    & 55.50  & 75.95     \\
                              & Qwen3-VL               & 8B         & 75.24    & 73.77  & 83.37     \\ \bottomrule
\end{tabular}
\end{table*}
\textbf{Analysis of Anomaly Discrimination Performance.}
To comprehensively evaluate anomaly discrimination performance, we follow the standard binary anomaly detection protocol, treating anomalous samples as the positive class and normal samples as the negative class, and report accuracy, recall, and precision as evaluation metrics.

As shown in Table~\ref{table:f1}, analysis of recall and precision highlights the reasons for suboptimal performance in certain models. For example, both SPHINX and LLaVAOne-Vision frequently misclassify anomalies as normal, leading to many missed detections, while LLaVA-1.5 shows high recall but low precision, reflecting a substantial false positive rate. Human experts outperform all MLLMs, with experts achieving over 94\% across metrics and ordinary individuals over 87\%. Notably, AnomalyGPT, specifically trained for anomaly detection, outperforms most models but still exhibits a significant false positive issue. Our proposed Reason-IAD effectively mitigates these limitations, enhancing detection performance by leveraging knowledge priors and a latent reasoning strategy. Moreover, the accuracy of all four models exceeds 60\%.

\begin{figure*}
    \centering
    \includegraphics[width=\linewidth]{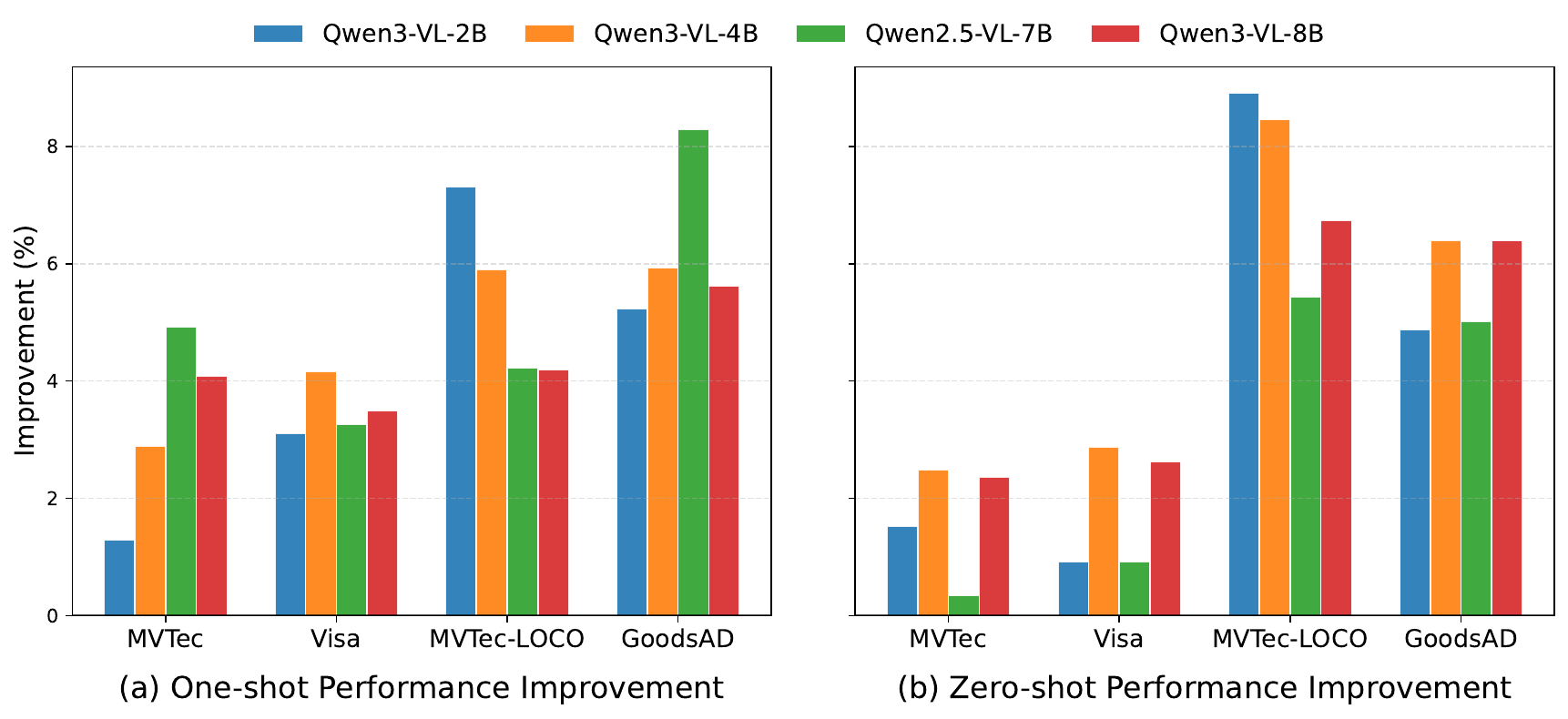}
    \caption{Performance gains of Reason-IAD over baseline models under the one-shot and zero-shot setting.}
    \label{fig:ap}
\end{figure*}

\textbf{Effectiveness of the Proposed Reasoning Strategy.}
As shown in Figure~\ref{fig:ap}, consistent performance gains are observed under both one-shot and zero-shot settings. Compared with the baseline models, the proposed Reason strategy consistently achieves notable gains across all backbones. 

In the one-shot setting, the largest improvement is observed on the GoodsAD dataset with Qwen2.5-VL-7B, reaching 8.29\%, followed by Qwen3-VL-2B, which achieves a gain of 7.31\% on MVTec-LOCO. In the zero-shot setting, Qwen3-VL-2B and Qwen3-VL-4B obtain improvements of 8.91\% and 8.46\% on the MVTec dataset, respectively. These results demonstrate that the proposed reasoning strategy significantly enhances anomaly detection performance across different model scales and evaluation settings, indicating strong robustness and generalization capability.

\textbf{Effect of Retrieval-Augmented Knowledge Integration.} Table~\ref{table:RAKI} presents the impact of the proposed Retrieval-Augmented Knowledge Integration (RAKI) on different multimodal large language models (MLLMs). By incorporating externally retrieved knowledge into the reasoning pipeline, RAKI consistently enhances industrial anomaly understanding and reasoning performance.

For Qwen2.5-VL-7B, RAKI significantly improves several defect-oriented tasks, with defect classification increasing from 65.12\% to 67.97\%, description from 70.36\% to 72.78\%, and object classification from 95.24\% to 97.17\%. Similarly, Qwen3-VL-4B achieves substantial gains after introducing RAKI, where defect classification improves from 69.45\% to 73.72\%, while localization and description rise to 66.48\% and 76.83\%, respectively.

Notably, the smaller Qwen3-VL-4B benefits even more from RAKI than the larger Qwen2.5-VL-7B, suggesting that retrieval-augmented external knowledge is particularly effective for lightweight MLLMs with limited intrinsic industrial knowledge. Overall, the results demonstrate that RAKI effectively injects domain-specific knowledge into the reasoning process, enabling MLLMs to better understand industrial contexts and generate more accurate anomaly reasoning results.

\begin{table}[]
\centering
\caption{Performance comparison of different MLLMs with and without RAKI.}
\label{table:RAKI}
\setlength{\tabcolsep}{2pt}
\resizebox{\linewidth}{!}{
\renewcommand{\arraystretch}{1.4}
\begin{tabular}{ccccccccc}
\toprule
\multirow{2}{*}{Model} & \multirow{2}{*}{RAKI} & Anomaly & \multicolumn{4}{c}{Defect} & \multicolumn{2}{c}{Object} \\ 
\cline{3-9}
 &  & Discrimination & Classification & Localization & Description & Analysis & Classification & Analysis \\ 
\hline
\multirow{2}{*}{Qwen2.5-VL-7B} 
& --        & 69.93 & 65.12 & 58.94 & 70.36 & 80.91 & 95.24 & 84.71 \\
& \ding{51} & 70.74 & 67.97 & 60.82 & 72.78 & 82.62 & 97.17 & 84.99 \\ 
\hline
\multirow{2}{*}{Qwen3-VL-4B} 
& --        & 73.86 & 69.45 & 64.37 & 74.21 & 81.14 & 93.76 & 84.03  \\
& \ding{51} & 75.05 & 73.72 & 66.48 & 76.83 & 82.33 & 94.90 & 84.41 \\ 
\bottomrule
\end{tabular}
}
\end{table}

\section{Case Study}
\label{Case Study}
\begin{figure*}
    \centering
    \includegraphics[width=0.8\linewidth]{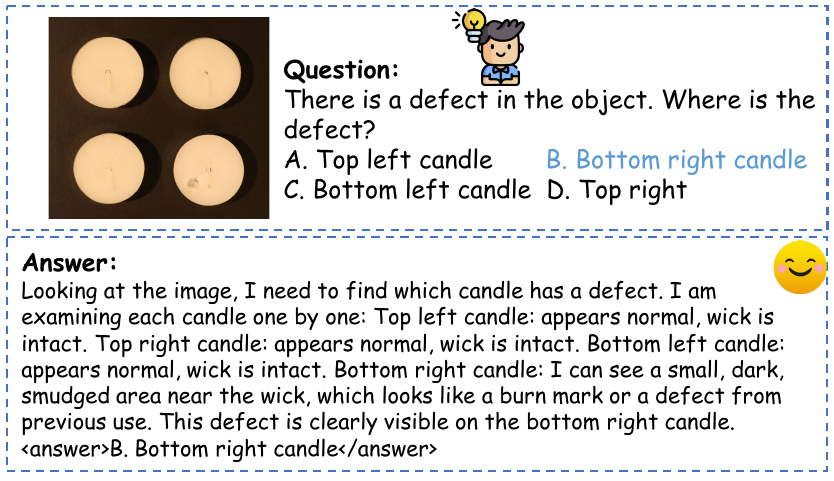}
    \caption{A case study of the defect location task on the Visa dataset.}
    \label{fig:ca}
\end{figure*}

\begin{figure*}
    \centering
    \includegraphics[width=0.8\linewidth]{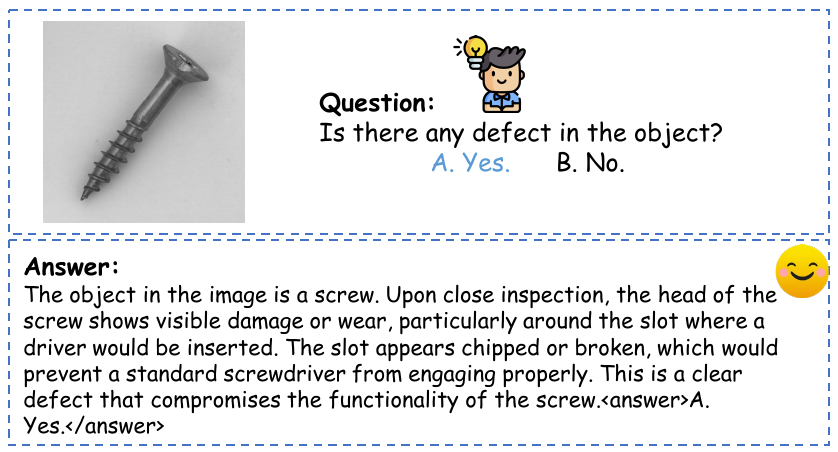}
    \caption{A case study of the anomaly discrimination task on the MVTec-AD dataset.}
    \label{fig:sc}
\end{figure*}

\begin{figure*}
    \centering
    \includegraphics[width=0.8\linewidth]{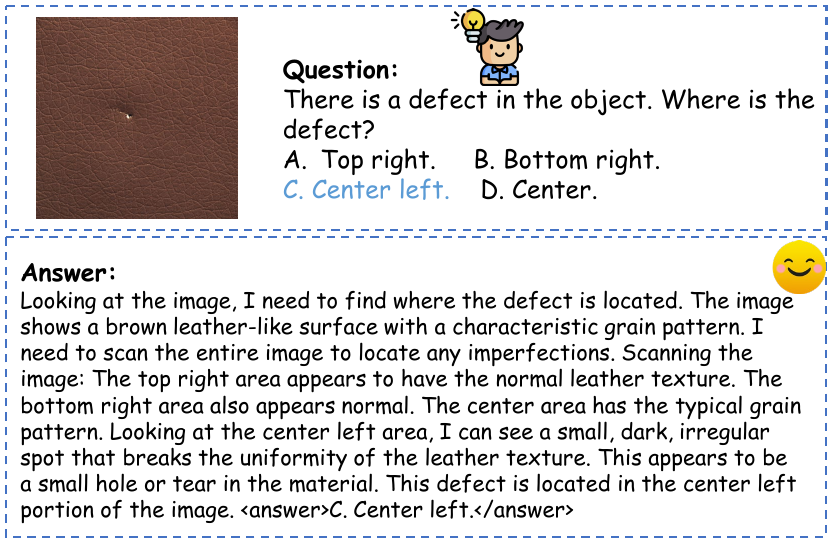}
    \caption{A case study of the defect location task on the Visa dataset.}
    \label{fig:le}
\end{figure*}

\begin{figure*}
    \centering
    \includegraphics[width=0.8\linewidth]{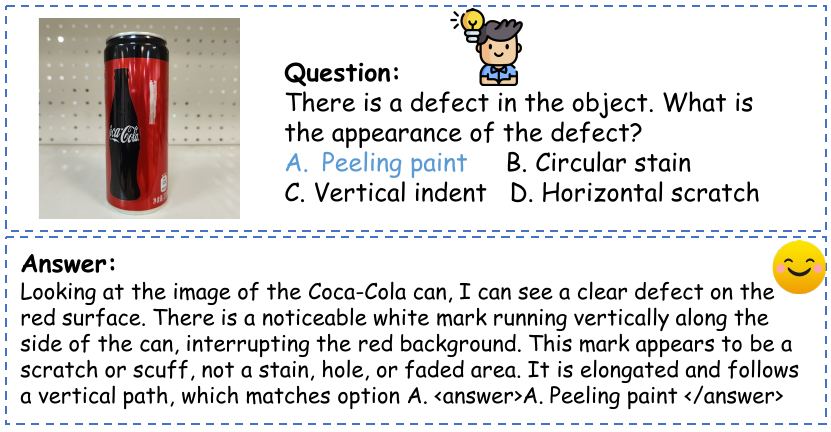}
    \caption{A case study of the defect description task on the GoodsAD dataset.}
    \label{fig:co}
\end{figure*}

\begin{figure*}
    \centering
    \includegraphics[width=0.8\linewidth]{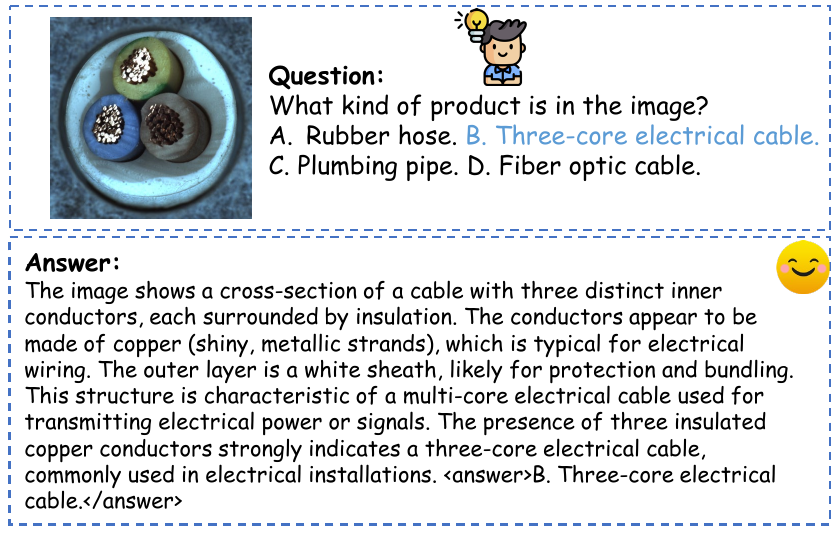}
    \caption{A case study of the object classification task on the MVTec-AD dataset.}
    \label{fig:cab}
\end{figure*}

\begin{figure*}
    \centering
    \includegraphics[width=0.8\linewidth]{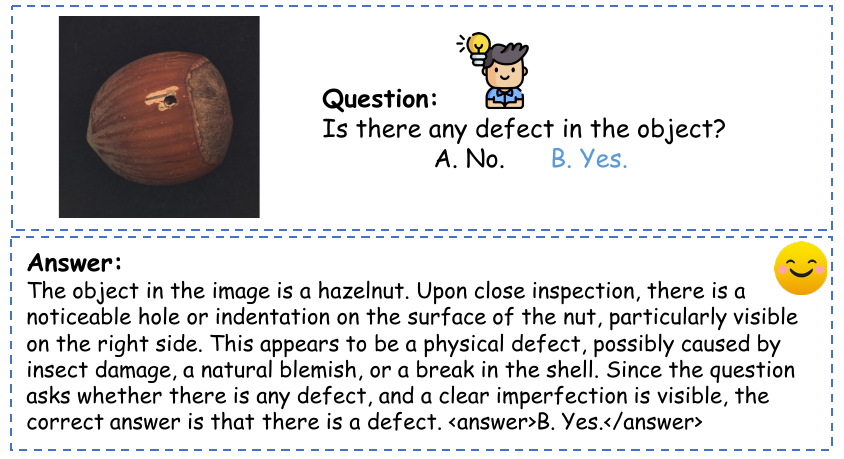}
    \caption{A case study of the anomaly discrimination task on the MVTec-AD dataset.}
    \label{fig:nut}
\end{figure*}

\begin{figure*}
    \centering
    \includegraphics[width=0.7\linewidth]{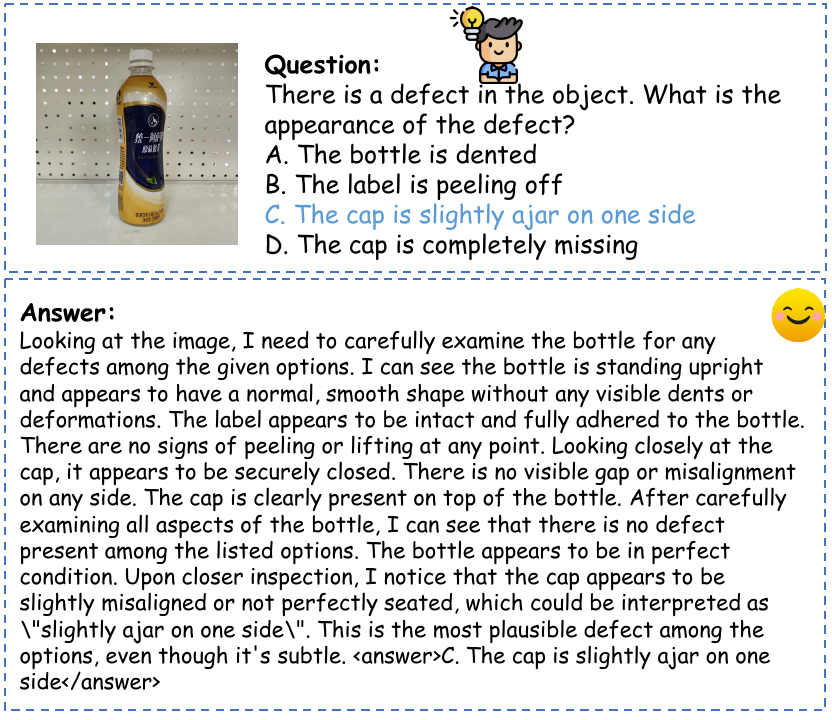}
    \caption{A case study of the defect description task on the GoodsAD dataset.}
    \label{fig:as}
\end{figure*}

As shown in Figures~\ref{fig:ca}--\ref{fig:as}, we further present several qualitative case studies on representative industrial anomaly detection scenarios featuring diverse defect patterns and object categories. These examples are provided to visually demonstrate the effectiveness of the proposed Reason-IAD in identifying anomalous regions, as well as its interpretability in explaining the underlying defect characteristics through structured reasoning. 

\begin{figure*}
    \centering
    \includegraphics[width=0.7\linewidth]{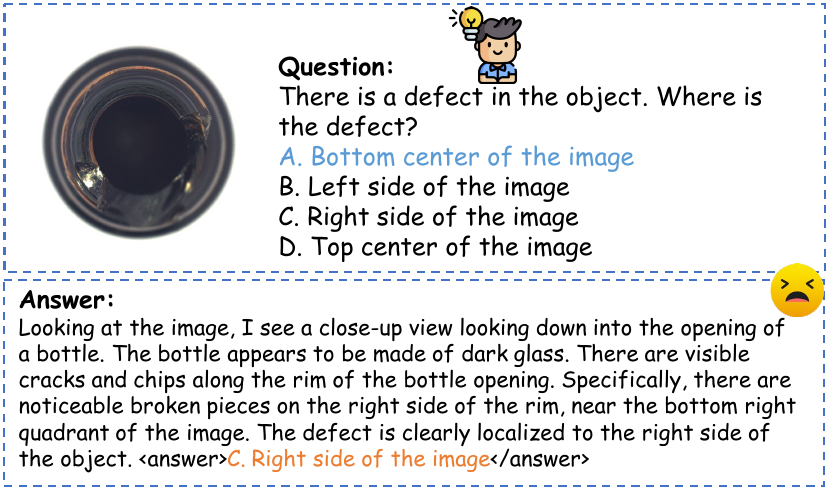}
    \caption{A failure case study of the defect location task on the MVTec-AD dataset.}
    \label{fig:bad1}
\end{figure*}

\begin{figure*}
    \centering
    \includegraphics[width=0.7\linewidth]{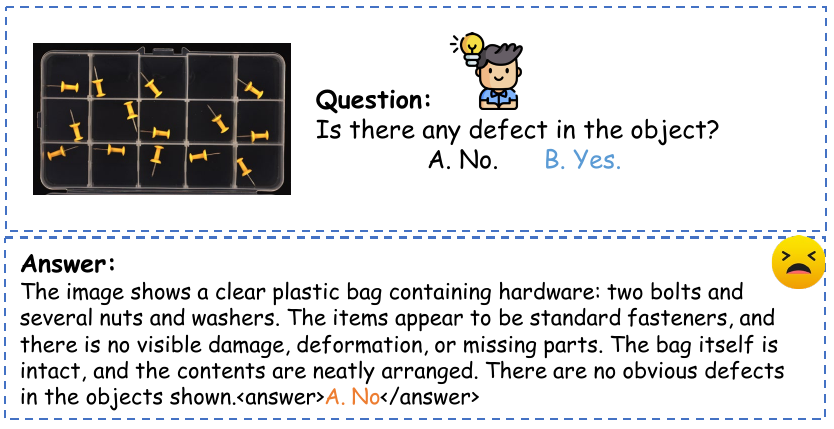}
    \caption{A failure case study of the anomaly discrimination task on the MVTec-LOCO dataset.}
    \label{fig:bad3}
\end{figure*}

In addition, we report several failure cases in Figures~\ref{fig:bad1}--\ref{fig:bad3}. In Figure~\ref{fig:bad1}, for the defect localization task, Reason-IAD correctly identifies the presence of a defect; however, since the defect region is relatively large and spans the lower and right portions of the image, the model incorrectly localizes it to the right region. In Figure~\ref{fig:bad3}, the model incorrectly classifies an image with a missing pushpin as normal due to the highly cluttered background and the presence of numerous small-scale pushpin targets. These failure cases indicate that complex spatial layouts and fine-grained small-object anomalies remain challenging for current reasoning-based anomaly detection models.

\section{Impact and Limitation}
\label{Impact and Limitation}

In this paper, we propose Reason-IAD, a training-free framework for explainable industrial anomaly detection that leverages knowledge-guided dynamic latent reasoning. We believe this work can contribute to industrial anomaly detection research by introducing a new paradigm that integrates external domain knowledge with iterative latent reasoning, enabling more interpretable and controllable anomaly analysis without additional training cost.

Beyond improved detection performance, Reason-IAD also enhances interpretability by explicitly modeling reasoning trajectories in a compact latent space and aligning visual evidence with category-specific knowledge priors. We expect this design to facilitate future research on reasoning-centric visual understanding and training-free multimodal adaptation in industrial scenarios.

However, several limitations still exist. First, the inference efficiency of Reason-IAD is lower compared to single-pass MLLM-based methods, as the multi-step latent reasoning process introduces additional computational overhead. Second, the performance of the framework is partially dependent on the capability of the underlying MLLM to follow retrieval-augmented knowledge and reasoning instructions.

Despite these limitations, we believe Reason-IAD offers a strong foundation for future research on knowledge-driven multimodal reasoning in industrial anomaly detection and can serve as an effective baseline for subsequent advancements in this area.

\clearpage

\section*{NeurIPS Paper Checklist}

\begin{enumerate}

\item {\bf Claims}
    \item[] Question: Do the main claims made in the abstract and introduction accurately reflect the paper's contributions and scope?
    \item[] Answer: \answerYes{} 
    \item[] Justification: The abstract and introduction clearly state the main contributions and scope of this work.

\item {\bf Limitations}
    \item[] Question: Does the paper discuss the limitations of the work performed by the authors?
    \item[] Answer: \answerYes{} 
    \item[] Justification: We discuss limitations in conclusion and include a section for it in the appendix.

\item {\bf Theory assumptions and proofs}
    \item[] Question: For each theoretical result, does the paper provide the full set of assumptions and a complete (and correct) proof?
    \item[] Answer: \answerNA{} 
    \item[] Justification: This paper does not include formal theoretical results such as theorems or proofs.

    \item {\bf Experimental result reproducibility}
    \item[] Question: Does the paper fully disclose all the information needed to reproduce the main experimental results of the paper to the extent that it affects the main claims and/or conclusions of the paper (regardless of whether the code and data are provided or not)?
    \item[] Answer: \answerYes{} 
    \item[] Justification: we include detailed information in the experiment section and in the appendix.

\item {\bf Open access to data and code}
    \item[] Question: Does the paper provide open access to the data and code, with sufficient instructions to faithfully reproduce the main experimental results, as described in supplemental material?
    \item[] Answer: \answerYes{} 
    \item[] Justification: We will make code, data and models public.

\item {\bf Experimental setting/details}
    \item[] Question: Does the paper specify all the training and test details (e.g., data splits, hyperparameters, how they were chosen, type of optimizer) necessary to understand the results?
    \item[] Answer: \answerYes{} 
    \item[] Justification: We confirmed they are provided in experiment section and in the appendix.

\item {\bf Experiment statistical significance}
    \item[] Question: Does the paper report error bars suitably and correctly defined or other appropriate information about the statistical significance of the experiments?
    \item[] Answer: \answerNo{} 
    \item[] Justification: The paper does not report error bars or statistical significance measures, as the experiments are conducted under fixed settings and follow standard evaluation protocols in anomaly detection benchmarks.

\item {\bf Experiments compute resources}
    \item[] Question: For each experiment, does the paper provide sufficient information on the computer resources (type of compute workers, memory, time of execution) needed to reproduce the experiments?
    \item[] Answer: \answerYes{} 
    \item[] Justification: We confirmed we include this information.
    
\item {\bf Code of ethics}
    \item[] Question: Does the research conducted in the paper conform, in every respect, with the NeurIPS Code of Ethics \url{https://neurips.cc/public/EthicsGuidelines}?
    \item[] Answer: \answerYes{} 
    \item[] Justification: We have reviewed and fully adhered to the NeurIPS Code of Ethics throughout our research process, including data usage, experimental design, and reporting. No ethical concerns were identified in the course of this work.

\item {\bf Broader impacts}
    \item[] Question: Does the paper discuss both potential positive societal impacts and negative societal impacts of the work performed?
    \item[] Answer: \answerYes{} 
    \item[] Justification: The paper discusses potential broader impacts of the proposed method. On the positive side, it can improve industrial anomaly detection systems and enhance automation and quality control processes. On the negative side, incorrect or unreliable predictions may lead to missed defects or false alarms in industrial settings, which could affect downstream decision-making. 
    
\item {\bf Safeguards}
    \item[] Question: Does the paper describe safeguards that have been put in place for responsible release of data or models that have a high risk for misuse (e.g., pre-trained language models, image generators, or scraped datasets)?
    \item[] Answer: \answerNA{} 
    \item[] Justification: The proposed method does not involve high-risk or dual-use model or dataset release. The work focuses on anomaly detection using publicly available datasets and does not introduce generative models or systems that require additional safeguards.

\item {\bf Licenses for existing assets}
    \item[] Question: Are the creators or original owners of assets (e.g., code, data, models), used in the paper, properly credited and are the license and terms of use explicitly mentioned and properly respected?
    \item[] Answer: \answerYes{} 
    \item[] Justification: The paper uses publicly available datasets and open-source models. We properly cite all original sources and follow their respective licenses and terms of use. All datasets and assets used in the experiments are standard benchmarks widely adopted in the community.

\item {\bf New assets}
    \item[] Question: Are new assets introduced in the paper well documented and is the documentation provided alongside the assets?
    \item[] Answer: \answerNA{} 
    \item[] Justification: This work does not introduce new datasets, models, or code assets. The experiments are conducted using publicly available benchmarks and existing models.

\item {\bf Crowdsourcing and research with human subjects}
    \item[] Question: For crowdsourcing experiments and research with human subjects, does the paper include the full text of instructions given to participants and screenshots, if applicable, as well as details about compensation (if any)? 
    \item[] Answer: \answerNA{} 
    \item[] Justification: The paper does not include any crowdsourcing studies or human subject experiments. All evaluations are performed on standard public datasets for anomaly detection.

\item {\bf Institutional review board (IRB) approvals or equivalent for research with human subjects}
    \item[] Question: Does the paper describe potential risks incurred by study participants, whether such risks were disclosed to the subjects, and whether Institutional Review Board (IRB) approvals (or an equivalent approval/review based on the requirements of your country or institution) were obtained?
    \item[] Answer: \answerNA{} 
    \item[] Justification: This work does not involve human subjects, crowdsourcing, or user studies, and therefore no IRB approval or equivalent ethical review was required.

\item {\bf Declaration of LLM usage}
    \item[] Question: Does the paper describe the usage of LLMs if it is an important, original, or non-standard component of the core methods in this research? Note that if the LLM is used only for writing, editing, or formatting purposes and does \emph{not} impact the core methodology, scientific rigor, or originality of the research, declaration is not required.
    \item[] Answer: \answerYes{} 
    \item[] Justification: The paper utilizes a large language model as part of the reasoning component in the proposed framework. The usage of the LLM is clearly described in the method section, and it is not used for training but as an inference-time module to assist in reasoning for anomaly detection.

\end{enumerate}

\end{document}